\newif\ifdoubleblind
\newif\iflackofspace

\documentclass{article}

\usepackage{natbib}

\usepackage{microtype}
\usepackage{graphicx}
\usepackage{subfigure}
\usepackage{amsmath,amssymb,amsfonts}
\usepackage{algorithmic}
\usepackage{xcolor}
\usepackage{csquotes}
\usepackage{pifont}
\usepackage{enumitem}
\usepackage{tabularx}
\usepackage{multirow}
\usepackage[hidelinks]{hyperref}
\usepackage[capitalise]{cleveref}

\def\BibTeX{{\rm B\kern-.05em{\sc i\kern-.025em b}\kern-.08em
    T\kern-.1667em\lower.7ex\hbox{E}\kern-.125emX}}

\usepackage[ruled,vlined,linesnumbered]{algorithm2e}

\setlist[itemize]{noitemsep, topsep=0pt}

\usepackage{textcomp}

\setlist[itemize]{noitemsep, topsep=0pt}
\usepackage{booktabs}

\usepackage{framed}
\usepackage[most]{tcolorbox}

\tcbset {
  enhanced jigsaw,%
  base/.style={
    arc=0mm, 
    bottomtitle=0.5mm,
    boxrule=0mm,
    colbacktitle=black!10!white, 
    coltitle=black, 
    fonttitle=\bfseries, 
    left=2.5mm,
    leftrule=1mm,
    right=3.5mm,
    title={#1},
    toptitle=0.75mm, 
  }
}

\definecolor{brandblue}{rgb}{0.34, 0.7, 1}
\newtcolorbox{mainbox}[1]{
  colframe=brandblue, 
  base={#1}
}

\newtcolorbox{subbox}[1]{
  colframe=black!30!white,
  base={#1}
}

\ifdoubleblind
\usepackage{mlsys2025}
\else
\usepackage[accepted]{mlsys2025}
\fi

\mlsystitlerunning{Automated Algorithm Design for Auto-Tuning Optimizers}

\usepackage[font=footnotesize,labelfont=footnotesize,skip=2pt]{caption}
\usepackage[font=scriptsize]{subcaption}
\usepackage{enumitem}
\setlist[itemize]{noitemsep, topsep=0pt}
\usepackage{booktabs}

\usepackage{parskip}
\begin{document}

\twocolumn[
\mlsystitle{Automated Algorithm Design for Auto-Tuning Optimizers}

\begin{mlsysauthorlist}
\mlsysauthor{Floris-Jan Willemsen}{liacs}
\mlsysauthor{Niki van Stein}{liacs}
\mlsysauthor{Ben van Werkhoven}{liacs}
\end{mlsysauthorlist}

\mlsysaffiliation{liacs}{LIACS, Leiden University, Leiden, the Netherlands}

\mlsyscorrespondingauthor{Floris-Jan Willemsen}{f.j.willemsen@liacs.leidenuniv.nl}

\mlsyskeywords{Auto-tuning, Optimization Algorithms, Large Language Models, High-performance Computing}

\vskip 0.3in

\begin{abstract}
\setlength{\parskip}{10pt}
Automatic performance tuning (auto-tuning) is essential for optimizing high-performance applications, where vast and irregular search spaces make manual exploration infeasible.
While auto-tuners traditionally rely on classical approaches such as evolutionary, annealing, or surrogate-based optimizers, designing algorithms that efficiently find near-optimal configurations robustly across diverse tasks is challenging.
We propose a new paradigm: using large language models (LLMs) to automatically generate optimization algorithms tailored to auto-tuning problems.
We introduce a framework that prompts LLMs with problem descriptions and search space characteristics to synthesize, test, and iteratively refine specialized optimizers.
These generated algorithms are evaluated on four real-world auto-tuning applications across six hardware platforms and compared against the state-of-the-art in two contemporary auto-tuning frameworks. 
The evaluation demonstrates that providing additional application- and search space-specific information in the generation stage results in an average performance improvement of 30.7\% and 14.6\%, respectively. 
In addition, our results show that LLM-generated optimizers can rival, and in various cases outperform, existing human-designed algorithms, with our best-performing generated optimization algorithms achieving an average 72.4\% improvement over state-of-the-art optimizers for auto-tuning.

\end{abstract}
]
\setlength{\parskip}{0pt}

\printAffiliationsAndNotice{}  %

\thispagestyle{plain}
\pagestyle{plain}

\section{Introduction}
\label{sec:introduction}

Automatic performance tuning, or auto-tuning, is an essential technique in high-performance computing (HPC) workloads such as scientific simulations and AI, enabling applications to deliver better performance and energy efficiency across diverse hardware architectures and input sizes~\cite{balaprakash2017autotuning, lessonsLearnedGPU2020, hijma2023optimization}. 
By systematically exploring a vast design space of functionally-equivalent program variants spanning thread layouts, data layouts, loop transformations, and a variety of other parameters, auto-tuners optimize the performance of software towards a specific hardware architecture~\cite{CLTune, KTT, ansel_opentuner_2014}. 
A central challenge in auto-tuning lies in efficiently navigating these large, noisy, irregular search spaces~\cite{willemsen2025SearchSpaceConstruction} as exhaustive exploration is infeasible~\cite{pruningRyoo, scloccoAutoTuningDedispersionManyCore2014}. Hence, the effectiveness of an auto-tuner critically depends on its optimization algorithms.
Classical approaches, including Simulated Annealing, Genetic Algorithms, and Particle Swarm Optimization, have been typically used in this domain, but require careful hyperparameter tuning~\cite{willemsenTuningTheTuner2025} and are not designed with the search space characteristics of auto-tuning in mind~\cite{willemsenBayesianOptimizationAutotuning2021}.
Meanwhile, recent advances in large language models (LLMs) have demonstrated remarkable capabilities in code generation, algorithm synthesis, and problem-solving across a wide range of domains~\cite{chenCodex2021evaluatinglargelanguagemodels, AlphaCodeLi2022}. These capabilities raise the question: Can LLMs be leveraged to automatically generate effective optimization algorithms for auto-tuning, benefitting from problem-specifics?
Such an approach could enable the rapid design of problem-specific optimizers without the need for expert-crafted heuristics, unlocking new efficiency levels in auto-tuning.

In this paper, we explore this novel direction by investigating the use of LLMs to generate optimization algorithms for auto-tuning. We integrate the auto-tuning framework Kernel Tuner~\cite{vanwerkhovenKernelTunerSearchoptimizing2019} and the LLaMEA framework, which combines LLMs with an evolutionary algorithm to automatically generate metaheuristics~\cite{vanstein2025llamealargelanguagemodel}.
We emphasize that only the optimization algorithm is generated, not the underlying applications to be auto-tuned, nor program variants. If a generated algorithm does not run or optimize correctly, it will not survive the selection phase of the evolutionary algorithm, ensuring only high-quality algorithms are produced. Our contribution therefore lies in improving search efficiency and robustness within a fixed, user-defined search space, rather than redefining kernel design or expanding the search space itself. 
Specifically, we evaluate whether LLM-generated algorithms can match or exceed the performance of human-designed optimization methods across a diverse benchmark suite, and whether such algorithms benefit from problem-specific information. 
Our work is, to the best of our knowledge, the first to systematically study the application of automated algorithm design in the context of auto-tuning.
We make the following contributions:
\begin{itemize}
\item We introduce the first closed-loop framework that evolves LLM-generated optimizers under formal auto-tuning evaluation metrics.
\item We evaluate LLM-generated optimizers against widely used classical algorithms across a representative set of real-world applications and architectures on both quality and efficiency.
\item We find that LLM-driven optimizer synthesis can outperform human-engineered search strategies even when decoupled from the tuning target.
\ifdoubleblind
\else
\item The best-performing generated algorithms are incorporated into the Kernel Tuner framework, benefiting users. 
\item All code and data are available in a GitHub repository\footnote{\tiny\url{https://github.com/XAI-liacs/BLADE/tree/paper/auto-tuning}}.
\fi
\end{itemize}

The remainder of this paper is structured as follows.
\Cref{sec:related_work} reviews related research on optimization methods and recent advances in LLM-based algorithm synthesis.
\Cref{sec:implementation} details our approach for generating and evaluating optimization algorithms with LLMs.
\Cref{sec:evaluation} presents a comprehensive evaluation.
Finally, \Cref{sec:conclusion_futurework} concludes. %

\section{Related Work}
\label{sec:related_work}

Auto-tuning is established as a key technique for optimizing performance-critical applications on modern heterogeneous systems~\cite{balaprakash2017autotuning}. 
ATLAS~\cite{whaleyAutomatedEmpiricalOptimizations2001} pioneered empirical auto-tuning by generating and benchmarking parameterized versions of dense linear algebra kernels, achieving performance portability across architectures. 
FFTW~\cite{fftw1998} introduced an adaptive planner that automatically searches among composition strategies for fast Fourier transforms. 
A number of frameworks have been developed over the years to generalize these application-specific approaches to arbitrary applications. 
For example, Active Harmony~\cite{ActiveHarmony} and Orio~\cite{Orio} extended auto-tuning to runtime adaptation and annotation-based empirical tuning, respectively. 
Open Tuner~\cite{ansel_opentuner_2014} is an extensible framework for application-level auto-tuning. 
CLTune~\cite{CLTune} is a generic auto-tuner for OpenCL kernels.
Traditionally, numerical optimization methods and metaheuristics have been widely used in these approaches, including Nelder-Mead, Genetic Algorithms (GA), Simulated Annealing (SA), Particle Swarm Optimization (PSO)~\cite{ActiveHarmony,Orio,vanwerkhovenKernelTunerSearchoptimizing2019,schoonhovenBenchmarkingOptimizationAlgorithms2022,ansel_opentuner_2014,CLTune}. 
Although effective in many settings, these algorithms generally require many empirical evaluations and careful hyperparameter choices to achieve good performance across diverse auto-tuning tasks~\cite{willemsenTuningTheTuner2025}.

\subsection{Machine Learning and Hybrid approaches}

Beyond traditional optimization algorithms, several auto-tuning approaches have explored alternative strategies. 
Multi-armed bandits and ensemble methods, as used in OpenTuner~\cite{ansel_opentuner_2014} and ATF~\cite{raschATFGenericAutoTuning2017}/PyATF~\cite{pyATF}, dynamically allocate budget across different optimizers, leveraging their strengths on different problem classes. 
Machine learning techniques have also been applied: predictive modeling approaches such as GPTune~\cite{liuGPTuneMultitaskLearning2021}, ytopt~\cite{wuYtoptAutotuningScientific2024}, and AUMA~\cite{AUMA} use regression models or Gaussian processes to guide the search towards promising configurations. 
KTT uses a machine-learning-based optimization method that incorporates performance counter data collected by a profiler~\cite{filipovicUsingHardwarePerformance2022}. Bayesian optimization frameworks specialized for high-performance computing, such as HyperMapper~\cite{nardiHyperMapperPracticalDesign2019}, further demonstrated the potential of learning-based auto-tuning. 
These methods show that performance can be improved not only by choosing better traditional optimizers, but also through hybrid, model-based, or domain-specific techniques~\cite{czarnulAutotuningMethodologyConfiguration2019,willemsenBayesianOptimizationAutotuning2021,wuAutotuningPolyBenchBenchmarks2020}.

\subsection{Meta-optimization techniques}

Recent work has begun to focus on meta-optimization, i.e., improving the auto-tuning process itself. 
\citet{willemsen2025SearchSpaceConstruction} studied how constraints shape the construction of search spaces in auto-tuning, while \citet{willemsenTuningTheTuner2025} introduced hyperparameter optimization for tuning algorithms within auto-tuners, demonstrating substantial performance gains. 
Constraint-aware optimization~\cite{willemsenExploringApplicationConstrained2025} further improves efficiency by preventing wasted evaluations on invalid configurations. Similarly, BaCO~\cite{hellstenBaCOFastPortable2024} automatically learns so-called {\em hidden constraints} to exclude configurations that turn out to be infeasible during compilation or at run time.
These contributions highlight the importance of adapting optimization algorithms to the unique challenges of auto-tuning.

\subsection{Automated Algorithm Design}

In parallel, the rise of large language models (LLMs) has inspired their application in code generation, software engineering, and algorithm design.
This opened up a new line of research, where LLMs are employed as generative engines for algorithms themselves. FunSearch \cite{funsearch} demonstrated that LLMs can discover competitive search procedures for combinatorial problems by generating functional code that is iteratively refined. Similarly, the Evolution of Heuristics (EoH) framework \cite{eoh} and ReEvo \cite{reevo} explore the automated design of heuristics and metaheuristics through evolutionary search guided by LLMs. Furthermore, LLaMEA (Large Language Model Evolutionary Algorithm,~\citet{vanstein2025llamealargelanguagemodel}) automatically evolves complete metaheuristic algorithms, showing that LLMs can propose novel optimization algorithms that are subsequently evaluated and selected in an evolutionary loop. Taken together, these works suggest a paradigm shift where optimizers are no longer hand-crafted but instead automatically generated, adapted, and evolved. An approach particularly promising for the domain-specific, large, noisy, irregular search spaces in auto-tuning.

While LLMs have been shown to be capable of generating competitive algorithms for classical continuous and combinatorial black-box optimization problems \cite{vanstein2025llamealargelanguagemodel,llameahpo}, their use in the domain of auto-tuning remains unexplored. 
Our work is the first to investigate LLMs as generative components for optimization algorithms within auto-tuning frameworks. 
Unlike most prior work, we do not rely on human-designed algorithm templates or hyperparameter tuning alone, but instead allow LLMs to propose entirely new optimization strategies, which are then evaluated under the rigorous autotuning methodology~\cite{methodologyPaper}. 

\section{Design and Implementation}
\label{sec:implementation}

\begin{figure*}
    \centering
    \includegraphics[width=0.9\linewidth]{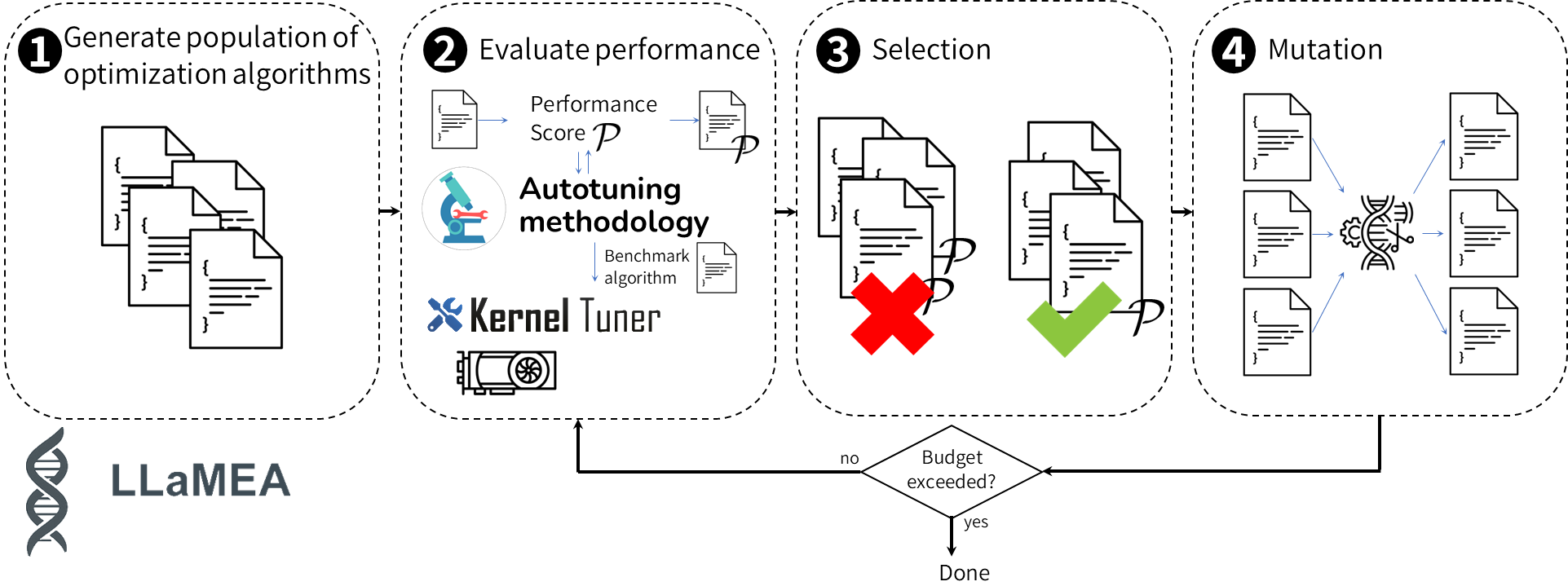}
    \caption{Overview of our designed system, showing how LLaMEA generates optimization algorithms, which are evaluated using the autotuning methodology, which tests the performance of the optimization algorithm in Kernel Tuner.}
    \label{fig:architecture}
\end{figure*}

This section describes the design and implementation of our approach. \Cref{fig:architecture} shows an overview of the design of our system to automatically generate optimization algorithms for auto-tuning. 
The LLaMEA search process is guided by an evolutionary algorithm (EA) that acts as a meta-strategy:
\begin{enumerate}
    \item A population of optimization algorithms is initialized with candidates generated by the LLM. In our case, LLaMEA starts with $4$ parent solutions.
    \item Each candidate algorithm is evaluated within Kernel Tuner using the autotuning methodology performance score $\mathcal{P}$ on the training set.
    \item Algorithms with higher scores are selected for reproduction, while poorly performing ones are discarded.
    \item A variety of LLM-driven mutation operators generate new candidate algorithms (in this case $12$), including diversity-focused mutation operators.
\end{enumerate}
Steps 2-4 are repeated for a fixed number of generations or until the evaluation budget is exhausted.

This design ensures that the LLM plays a creative but non-critical role: it proposes optimization algorithms, while the EA-based selection mechanism ensures only high-performing candidates survive. Because evaluation is entirely based on measured performance in Kernel Tuner, the process is robust to errors or inefficiencies in LLM-generated code. In addition, LLaMEA handles time-outs and errors in generated code by feeding back the stacktraces as additional context for the LLM, which allows for self-debugging when needed. %

The rest of this section explains the individual components and details how these are integrated to work in concert.
We first introduce Kernel Tuner as the target auto-tuning framework in \cref{subsec:kernel-tuner}, followed by \cref{subsec:llamea} with an overview of how we have used the LLaMEA system for LLM-assisted automated algorithm design. Finally, we explain how these components are integrated in \cref{subsec:integration}: LLaMEA generates Kernel Tuner-compatible optimization algorithms that are evaluated with a performance score, while an evolutionary algorithm guides the evolutionary search over candidate algorithms.

\subsection{Kernel Tuner}
\label{subsec:kernel-tuner}

Kernel Tuner is a Python-based, open-source auto-tuning framework designed for optimizing computational kernels~\cite{vanwerkhovenKernelTunerSearchoptimizing2019}. It supports multiple programming backends, including CUDA, HIP, OpenCL, and OpenACC, C++ and Fortran, and is widely adopted for tuning applications across different architectures and objectives, such as execution time and energy efficiency.

\begin{figure}
    \centering
    \includegraphics[width=0.95\linewidth]{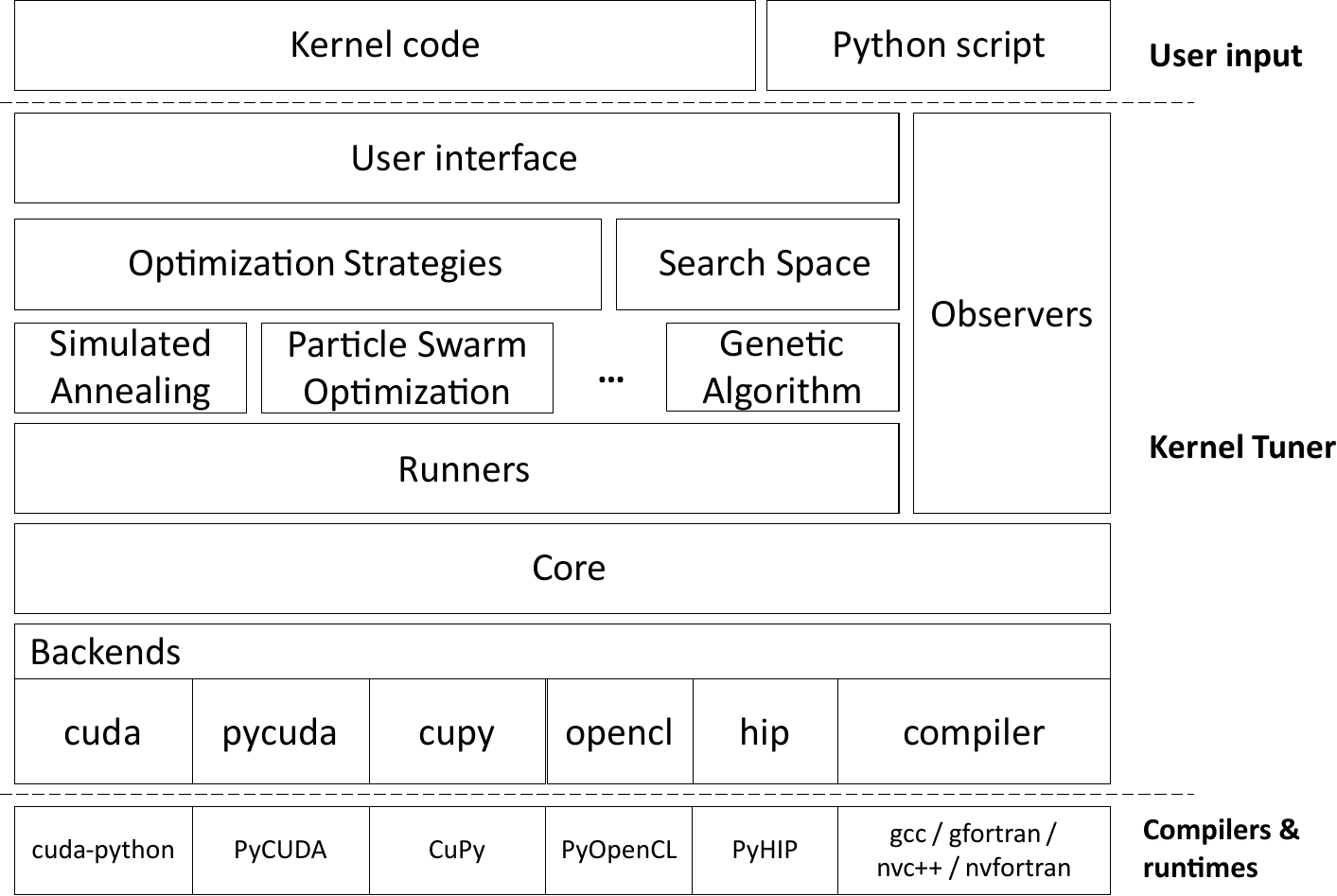}
    \caption{Overview of the software architecture of Kernel Tuner.}
    \label{fig:kt-arch}
\end{figure}

\cref{fig:kt-arch} shows the overview of the software architecture of Kernel Tuner.
Users provide Kernel Tuner with a kernel implementation and a Python script that specifies the input data, a description of the tunable parameters, and possibly any constraints. At the start of the tuning process, Kernel Tuner constructs a search space $\mathcal{X}$ consisting of all valid configurations. The optimization algorithm, called an optimization strategy in Kernel Tuner, iteratively selects candidate solutions $x \in \mathcal{X}$ to be compiled, executed, and measured. The evaluation metric, typically kernel runtime, determines the quality of the configuration. Kernel Tuner internally abstracts the different backends into a unified interface, allowing most of the tool to operate independently of the programming language and target hardware platform. 

More formally, auto-tuning involves optimizing an application or {\em kernel} $K_i$ on a target system $G_j$ for input data $I_k$ to maximize performance $f_{G_j,I_k}(K_i)$. 
The auto-tuner constructs a search space $\mathcal{X}$ by considering all tunable parameters and their valid values, subject to user-defined constraints~\cite{willemsen2025SearchSpaceConstruction}. 
The objective is to determine the optimal configuration, or more formally (assuming minimization), as follows: 
\vspace{-3mm}
\begin{equation} \label{eq:autotuning}
x^\star = \underset{x\in\mathcal{X}}{\text{arg min}} \, f_{G_j,I_k}(K_{i,x}).
\end{equation}

The evaluation of each configuration takes time, as it needs to be compiled and executed on the target system. 
Search spaces in auto-tuning are often large, discontinuous, non-convex, and irregular. 
These characteristics make them infeasible to search by hand and demand carefully chosen optimization algorithms. 
If the algorithm is not well adapted to the search space, the number of evaluations taken to find a near-optimal configuration becomes excessive, limiting the practical usability of the auto-tuner. We have extended Kernel Tuner with the capability to use any externally-defined optimization algorithm, not only those already included in the Kernel Tuner source code. To this end, we have implemented an \verb|OptAlg| wrapper class that can be used to wrap optimization algorithms in a format that Kernel Tuner supports.

\subsection{LLaMEA}
\label{subsec:llamea}

The Large Language Model Evolutionary Algorithm~\cite{vanstein2025llamealargelanguagemodel}, LLaMEA for short, is a system designed to leverage the generative capabilities of large language models for the synthesis of optimization algorithms. In our setting, the scope of the LLM is \emph{strictly limited to generating the optimization algorithm logic} during an algorithm design stage that is separate from the auto-tuning stage. This ensures that the LLM does not affect application kernels, data, execution, or numerical accuracy. Consequently, the correctness and reproducibility of results remain intact, as the kernel execution pipeline remains unchanged.
The LLM generates code snippets that define candidate optimization algorithms. The optimization algorithms specify how new configurations are sampled, selected, and evaluated within Kernel Tuner. If a generated algorithm is incorrect or contains implementation errors, it simply yields poor performance when evaluated, resulting in a low performance score (fitness). These weak algorithms are discarded during the evolutionary search process.
The initial population of candidate optimization algorithms is generated using the prompt template shown in \cref{fig:prompt-template}. 
The \verb|code format specification| briefly explains how to use Kernel Tuner's \verb|OptAlg| base class as well as the interface of the \verb|SearchSpace| object in Kernel Tuner (see \cref{fig:kt-arch}).
Optionally, we can insert information about the specific tuning problem at hand to allow the LLM to incorporate information on the tunable parameters, their possible values, and constraints. 
The \verb|minimum working code example| includes an example implementation of an empty optimization algorithm template, illustrating how to use the \verb|SearchSpace| object to: 1) generate an initial population, 2) retrieve neighbors of a particular configuration, and 3) repair any invalid configuration that violates constraints.
Finally, the \verb|output format specification| instructs the LLM to first print a one-line description followed by the code.
Complete prompts can be found in the provided repository.

\begin{figure}[tbp]
\begin{mainbox}{Task Prompt} 
    \scriptsize
    Your task is to design novel metaheuristic algorithms to solve kernel tuner problems (integer, variable dimension, constraint). \\
    \texttt{$<$code format specification$>$} \\
    \texttt{$<$OPTIONAL search space specification (json)$>$}\\
    An example code structure with helper functions is as follows:\\
    \texttt{$<$minimum working code example$>$} \\
    Give an excellent and novel heuristic algorithm to solve this task and also give it a one-line description, describing the main idea.\\
    \texttt{$<$output format specification$>$}
\end{mainbox}
\vspace{-5pt}
\caption{Prompt template used to generate initial candidate optimization algorithms.}
\label{fig:prompt-template}
\end{figure}

\begin{figure}[tbp]
\begin{mainbox}{Mutation Prompts} 
    \scriptsize
    \begin{itemize}
        \item Refine the strategy of the selected solution to improve it.
        \item Generate a new algorithm that is different from the algorithms you have tried before.
        \item Refine and simplify the selected algorithm to improve it.
    \end{itemize}
\end{mainbox}
\vspace{-5pt}
\caption{Mutation prompts used by LLaMEA for Kernel Tuner.}
\label{fig:prompt-mutation}
\end{figure}

LLaMEA employs an iterative evolutionary loop in which LLMs generate candidate metaheuristic algorithms in executable code, which are then autonomously evaluated to provide performance feedback. The framework supports mutation and selection strategies common in evolutionary computing algorithms with different solution population sizes.
Code evolution is implemented via different natural language-based mutation prompts (See \cref{fig:prompt-mutation}). These prompts are selected to give both exploration and exploitation behaviour in code space. LLaMEA for Kernel Tuner uses an elitism strategy with 4 parent solutions and 12 offspring solutions per iteration. Beyond automated algorithm discovery, LLaMEA offers capabilities such as in-the-loop hyperparameter tuning \cite{llameahpo}, where numerical tuning is delegated to hyperparameter optimization tools, allowing the LLM to focus on structural innovation. 

\subsection{Rating Automatically Generated Algorithms}
\label{subsec:integration}

Kernel Tuner includes more than 20 optimization algorithms~\cite{schoonhovenBenchmarkingOptimizationAlgorithms2022}, ranging from local search to population-based methods. However, large and irregular search spaces present in auto-tuning make algorithm design challenging. Ineffective optimizers waste evaluations on poor configurations, reducing tuning efficiency. Our approach addresses this challenge by automatically generating novel optimization algorithms tailored for auto-tuning and using Kernel Tuner-specific operations. 
Nevertheless, evaluating whether an optimization algorithm is actually well-suited for auto-tuning is challenging; as the large, irregular search spaces encountered in auto-tuning can be difficult to optimize, an optimizer that performs well on one search space may perform poorly on another. 
A robust, systematic approach to evaluation is thus needed to automatically generate suitable algorithms.

\ifdoubleblind
\else
In earlier work, we presented a community-driven methodology for evaluating optimization algorithms in auto-tuning~\cite{methodologyPaper}, which can be used to evaluate LLaMEA algorithms with Kernel Tuner. 
\fi
This methodology provides a systematic approach to comparing optimization algorithms across auto-tuning search spaces. 
It defines a performance score $\mathcal{P}$ that quantifies an optimization algorithm's performance over the passed time relative to a calculated baseline, typically random search, to have consistent, objective-independent, transparent, and comparable behavior across search spaces. 

On an individual search space, this approach first defines a budget and performance baseline adapted to the search space characteristics. 
The optimization algorithms to be compared are then executed multiple times to account for noise, after which the difference between the baseline and the average best performance of each optimization algorithm is compared at fixed, equidistant time intervals relative to the budget. 
The result is a smooth performance curve over time, %
instead of relying solely on final performance. %
By adapting the budget and baseline to reliable search space characteristics, we can compare and aggregate these performance curves across search spaces. 
We will thus use the mean of these aggregated performance curves as a performance score $\mathcal{P}$ of an optimization algorithm, where a higher score is better overall performance. 

More formally, for a given time sampling point $t$ the performance $\mathcal{P}_t$ of an optimization algorithm $\mathcal{F}$ can be computed:
\vspace{-0.2cm}
\begin{equation} \label{eq:performance_curve}
\mathcal{P}(\mathcal{F},G_j,K_i,I_k)_t = \frac{\mathcal{S}_{\text{baseline}}(t) - \mathcal{F}(G_j,K_i,I_k)_t}{S_{\text{baseline}}(t) - \mathcal{S}_{\text{opt}}}
\end{equation}
where $\mathcal{S}(G_j,K_i,I_k)$ are the search space characteristics given a target system $G_j$, kernel $K_i$, and input data $I_k$. 
Here, $\mathcal{F}_t$ is the best objective value found so far by the optimization algorithm, $\mathcal{S}_{\text{baseline}}(t)$ is the performance of the baseline method, and $\mathcal{S}_{\text{opt}}$ is the known optimal value in the search space. 
This yields $\mathcal{P}_t = 0$ when performance equals the baseline and $\mathcal{P}_t = 1$ when the optimum has been found. 

To avoid distorting the metric with large variations in search space difficulty, evaluations are limited to the time at which the baseline reaches a set cutoff percentile between the median and the optimum, typically somewhere around 95\%, referred to as the budget. 
With this method, the performance curves are relative to the same baseline and optimum, the cutoff provides a consistent reference point, and the sampling points are equidistant. 
The performance curves can thus be meaningfully aggregated from different search spaces by taking the mean score of all curves at each $t$, resulting in the aggregate performance curve over all search spaces for the optimization algorithm. 
The aggregate performance score is then obtained by averaging over the set of discrete time sampling points $\mathcal{T}$, or more formally:
\vspace{-1.7mm}
\begin{equation} \label{eq:performance_score}
\mathcal{P}(\mathcal{F},K,G,I) = \frac{1}{|\mathcal{T}|} \sum_{t \in \mathcal{T}} \frac{\displaystyle \sum_{K_i \in K} \sum_{G_j \in G} \sum_{I_k \in I} \mathcal{P}(\mathcal{F}, G_j, K_i, I_k)_t}{|K||G||I|}
\end{equation}
The kernels $K$, target systems $G$, and inputs $I$ are the collections of $K_i$, $G_j$, and $I_k$ of \cref{eq:autotuning} on which the LLaMEA algorithms are evaluated, hence referred to as the training set. 
This aggregate score enables robust comparison of optimization algorithms by capturing both the quality of the configurations found as well as the time taken to do so. 
As such, a difference when comparing two performance scores can indicate a difference in the quality of configurations found, the time taken to do so, or a combination of both.

\section{Evaluation}
\label{sec:evaluation}

In this section, we evaluate the effectiveness of large language model (LLM)-generated optimization algorithms as introduced in \cref{sec:implementation}. 
\Cref{subsec:evaluation_setup} describes the experimental setup. 
We then address our two main research questions: 
whether our LLM-generated algorithms benefit from incorporating additional problem-specific information (\cref{subsec:evaluation_results_problem_specificity}), 
and whether our generated algorithms can match or exceed the performance of established optimization methods for auto-tuning (\cref{subsec:evaluation_results_traditional}). 
In between, \cref{subsec:evaluation_algorithm_details} discusses the details of the two best generated algorithms.

\subsection{Experimental Setup}
\label{subsec:evaluation_setup}

The experimental setup consists of five sections: \cref{subsec:evaluation_setup_benchmark} detailing the benchmarks evaluated on, 
\cref{subsec:evaluation_setup_hardware} describing the hardware and execution, an account of the software and models in \cref{subsec:evaluation_setup_software}, an analysis of the costs in \cref{subsec:evaluation_setup_cost}, and an outline of the metrics used in \cref{subsec:evaluation_setup_metrics}.

\subsubsection{Benchmarks} \label{subsec:evaluation_setup_benchmark}

\begin{table}[tb]
    \centering
    \footnotesize
    \caption{Overview of the basic characteristics of the real-world applications.}
    \label{tab:searchspaces_real_world_overview}
    \begin{tabularx}{\linewidth}{l|X|X|X}
    \toprule
        \textbf{Name} & \textbf{Cartesian size} & \textbf{Constrained size} & \textbf{Dimensions} \\
    \midrule
        Dedispersion    & 22272     & 11130  & 8   \\
        2D Convolution  & 10240     & 4362   & 10  \\ 
        Hotspot         & 22200000  & 349853 & 11 \\
        GEMM            & 663552    & 116928 & 17  \\
    \bottomrule
    \end{tabularx}
\end{table}

We will evaluate our approach using the BAT benchmark suite~\cite{BenchmarkingSuiteKerneltuners} of auto-tunable GPU kernels. Specifically, we use the \textit{dedispersion}, \textit{convolution}, \textit{hotspot}, and \textit{GEMM} kernels. These benchmark kernels are examples of widely used real-world applications in astronomy, image processing, materials science, and linear algebra, respectively. 
The characteristics of these applications are shown in~\cref{tab:searchspaces_real_world_overview}. The Cartesian size is the size of the complete combinatorial space. %
The constrained size is the number of feasible solutions that remain after applying constraints. The number of dimensions equals the number of tunable parameters in the source code. 
We use these four applications as a benchmark throughout this evaluation. Besides diverse origins, they also bring diversity in their performance characteristics; e.g., dedispersion and hotspot are generally bandwidth-bound, while convolution and GEMM are generally compute-bound; we will explore these applications in more detail.

The \emph{Dedispersion} kernel in BAT originates from the AMBER pipeline for the detection of single-pulse astronomical transients~\cite{sclocco2020amber}. Dedispersion is a signal processing operation that compensates for the frequency-dependent time delay experienced by radio waves as they propagate through space. This delay, caused by dispersion in the interstellar medium, results in lower-frequency components of the signal arriving later than higher-frequency components that were emitted simultaneously. For a signal with the highest frequency $f_h$ received at time $t_x$, the delay $k$ for a lower frequency $f_i$ can be approximated by:
\vspace{-2mm}
\begin{equation*}
    k \approx 4150 \times DM \times \left( \frac{1}{f_i^2} - \frac{1}{f_h^2} \right),
\end{equation*}
where $DM$ is the dispersion measure, characterizing the integrated column density of free electrons between the source and the observer.
The kernel takes as input time-domain samples across many frequency channels and outputs dedispersed samples for a range of $DM$ values. It is parallelized such that each thread can process multiple time samples and dispersion measures in parallel, iterating over the frequency bands. For this study, we use parameters based on the ARTS survey on the Apertif telescope~\cite{van2022apertif}, which employs a sampling rate of 24.4 kHz, 2048 dispersion measures, and 1536 frequency channels.
Several tunable parameters influence the performance of this kernel on GPUs. These include the dimensions of the thread blocks, the amount of work assigned per thread in both the time and dispersion measure dimensions, and the strategy used to stride through the input samples. The kernel also allows partial loop unrolling over the frequency channels, where a factor of zero delegates this decision to the CUDA compiler. Additionally, the number of thread blocks per streaming multiprocessor can be controlled to optimize resource utilization. %

The \emph{Convolution} kernel, developed by~\citet{vanwerkhovenOptimizingConvolutionOperations2014} as an optimized and highly tunable GPU-accelerated library for 2D convolution operations, has become a widely used benchmark in autotuning research~\cite{CLTune,petrovicBenchmarkSetHighlyefficient2020,vanwerkhovenKernelTunerSearchoptimizing2019,schoonhovenBenchmarkingOptimizationAlgorithms2022}. 
A two-dimensional convolution is a fundamental operation in image processing and computer vision, where a filter mask is applied across an input image to compute a weighted sum of neighboring pixel values for each output element. Given an input image $I$ of size $(w \times h)$ and a convolution filter $F$ of size $(F_w \times F_h)$, the output image $O$ of size $((w-F_w) \times (h-F_h))$ is computed:
\vspace{-3mm}
\begin{equation}\nonumber
O(x,y) = \sum\limits_{j=0}^{F_h} \sum\limits_{i=0}^{F_w} I(x+i,y+j)\times F(i,j)
\end{equation}
The implementation in BAT exposes several tunable parameters that strongly influence GPU performance. The thread block dimensions in the $x$ and $y$ directions determine the granularity of parallelism and the number of threads per block. Each thread may compute multiple output elements along the $x$ and $y$ axes, depending on the selected tile sizes. A shared memory padding scheme can be enabled to mitigate bank conflicts when block sizes are not multiples of the number of memory banks, as described by~\citet{vanwerkhovenOptimizingConvolutionOperations2014}. Additionally, the kernel can optionally use the read-only cache to reduce global memory traffic when loading input elements.

The \emph{Hotspot} kernel, included in BAT and originating from the Rodinia Benchmark suite~\cite{cheRodiniaBenchmarkSuite2009}, is part of a thermal simulation application that estimates the temperature distribution of a processor. The simulation considers the processor's architectural floor plan, thermal resistance, ambient temperature, and simulated power currents. Through an iterative process, the kernel solves a set of differential equations to model heat dissipation over time. The inputs to the kernel are the power consumption and initial temperature values, and the output is a two-dimensional grid of average temperature values across the chip.
The performance of the Hotspot kernel is highly dependent on a number of tunable parameters that influence how work is distributed across GPU threads and how data is managed in memory. The thread block dimensions in the $x$ and $y$ directions define the number of threads per block, with between 32 and 1024 threads typically used. Each thread may compute one or more output elements in each dimension. The temporal tiling factor controls whether multiple iterations of the stencil operation are to be performed within a single kernel launch, improving data locality. %
Finally, shared memory usage for caching power input data can be enabled, and another parameter controls the number of thread blocks per SM to influence occupancy and register usage.

\emph{Generalized dense matrix-matrix multiplication (GEMM)} is a fundamental operation defined in the BLAS linear algebra specification and is extensively used across scientific computing, machine learning, and high-performance computing applications. 
It is also a common benchmark for GPU code optimization studies~\cite{CLTune,li2009note,pruningRyoo} due to its computational intensity and sensitivity to hardware-specific optimizations. 
In this evaluation, we use the GEMM kernel from CLBlast~\cite{nugterenCLBlast2018}, a tunable OpenCL BLAS library.
The GEMM operation performs the multiplication of two matrices, $A$ and $B$, and optionally adds a scaled version of an existing output matrix: $C = \alpha A \cdot B + \beta C$
where $\alpha$ and $\beta$ are scalar coefficients.
The CLBlast GEMM kernel exposes a range of tunable parameters that affect its performance on different GPU architectures. 
The tunable parameters control the workload assigned to each thread block in two dimensions, while another two parameters control the dimensions of the thread block itself. 
Loading matrix tiles into shared memory can be enabled or disabled for both matrix $A$ and $B$ individually. When enabled, two more parameters control shared memory level tile sizes.
Finally, two more parameters define the vector widths used for global memory transactions, enabling efficient vectorized loads and stores. 

These benchmark applications and their described parameters each form a flexible search space for performance tuning across different GPU architectures.

\subsubsection{Hardware and Execution} \label{subsec:evaluation_setup_hardware}
To obtain a diverse set of real-world auto-tuning cases for evaluation, we use the four auto-tuning applications described in \cref{subsec:evaluation_setup_benchmark} on a diverse set of six GPUs from the DAS-6~\cite{DASMediumScaleDistributedSystem} and LUMI~\cite{LUMIsupercomputer} supercomputers, resulting in 24 unique auto-tuning search spaces. 
The generation phase aims to discover robust heuristic strategies for a class of auto-tuning problems using a representative benchmark set, consistent with standard practice in optimizer design. 
For fair evaluation, the LLaMEA optimization algorithm feedback loop uses a training set of twelve search spaces resulting from the four applications on the AMD MI250X, Nvidia A100, and Nvidia A4000, and the test set consists of twelve search spaces from the four applications on the AMD W6600, AMD W7800, and Nvidia A6000. 
Even when evaluating the same kernels across GPUs, the resulting search spaces differ substantially due to hardware-specifics~\cite{luratiBringingAutoTuningHIP2024}, ensuring that performance reflects robustness across diverse tuning landscapes rather than memoization of specific configurations.

As a result, the generation cost is incurred only once, after which the cost is quickly amortized across many runs, as described in \cite{willemsenTuningTheTuner2025}. 
Optimizer evaluation is accelerated by leveraging these pre-exhaustively explored search spaces, allowing simulation rather than recurring recompilation and kernel execution. This reduces evaluation time for a candidate optimizer to seconds or minutes rather than hours or days. 
The evaluation of this work is conducted on the A4000 nodes of DAS6. 
The nodes in DAS-6 have a 24-core AMD EPYC-2 7402P CPU and 128 GB of RAM. 

\subsubsection{Software and Models} \label{subsec:evaluation_setup_software}
The OS used is Rocky Linux 8 with Linux kernel version 4.18, the Python version 3.11.7, as well as \href{https://github.com/KernelTuner/kernel_tuner/releases}{Kernel Tuner 1.3.0}, and PyATF 0.0.9.
The version of \href{https://github.com/XAI-liacs/LLaMEA/releases}{LLaMEA used is 1.0.6}, and GPT o4-mini with the default hyperparameters is used as the LLM. 
We initially experimented with \texttt{gemini-2.0-flash} and \texttt{o4-mini} models due to their cost-effectiveness, and although \texttt{gemini-2.0-flash} produced valid algorithms, their performance was generally inferior to those generated by \texttt{o4-mini}. 
There is a wide variety of other, more powerful models that are likely to outperform the models mentioned, yet with which it would not have been feasible to conduct these experiments due to cost. Our goal is to assess whether LLM-driven optimizer synthesis for auto-tuning is practically viable, rather than to maximize absolute performance using the largest available models. 
The fact that strong and competitive optimizers emerge even with relatively lightweight models demonstrates both the robustness of the approach and its accessibility to a broad set of users. 

\subsubsection{Cost of Optimizer Generation} \label{subsec:evaluation_setup_cost}
\begin{figure}[tb]
    \centering
    \includegraphics[width=\linewidth]{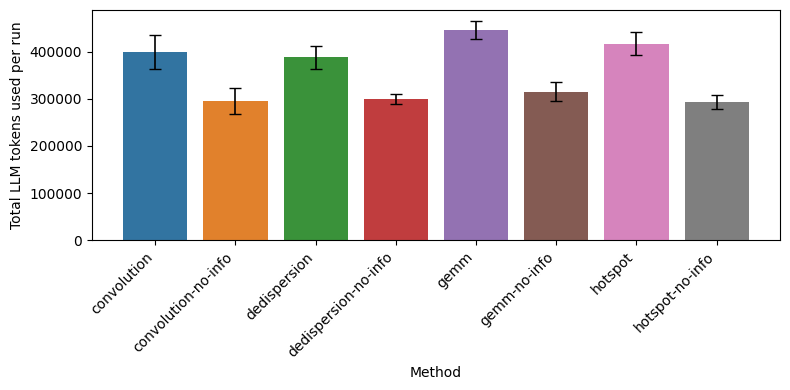}
    \caption{
    The total number of LLM tokens per generated optimizer, averaged over five runs. Error bars denote the standard deviation over the five runs.}
    \label{fig:evaluation_llm_cost_token_count_per_optimizer}
\end{figure}

We report the cost and practical limitations of the LLM-based optimizer generation process to provide a realistic view of its feasibility. Across all experiments, optimizer generation required 100 LLM calls per run, with 5 independent runs per kernel and method to obtain statistically meaningful results, resulting in a total of 4000 LLM calls. 
Out of the 5 independent run, the best-performing optimization algorithm was selected. 
The overview of the total token count for each generated optimizer is shown in \cref{fig:evaluation_llm_cost_token_count_per_optimizer}. 

During generation, each candidate optimizer is evaluated with a maximum wall-clock time of 5 minutes; candidates exceeding this limit are discarded. In practice, most evaluations complete well within this bound due to the use of pre-exhaustively explored search spaces, which avoid expensive compilation and execution.

We observe an approximate 25\% failure rate among generated optimizers, including invalid code, runtime errors, or exceeding the time limit. Failed candidates are simply discarded from the population. Given the evolutionary setup with 4 parents and 12 offspring candidates per generation, it is unlikely that all candidates fail simultaneously. In the rare case that this occurs, the LLM is provided with the corresponding stack traces and prompted to repair the implementation in the next iteration, which we find to be consistently effective in practice.

\subsubsection{Metrics} \label{subsec:evaluation_setup_metrics}
For performance comparison, we follow the auto-tuning methodology community standard~\cite{methodologyPaper} as detailed in \cref{subsec:integration}. 
To provide an intuition of this metric, it measures the area under the performance-versus-time curve, normalized to a baseline, and is intended to capture both (i) how quickly an optimizer approaches high-quality configurations and (ii) the final quality achieved within a fixed tuning budget. This reflects a core practical constraint in auto-tuning, where users care about reaching good objective values within a limited amount of time. 
More specifically, each optimization algorithm is evaluated using a \emph{performance score} $\mathcal{P}$ that measures the quality of configurations found over time relative to a statistically estimated \emph{random search baseline}. 
This aggregate metric captures both search efficiency and solution quality across different search spaces. 
A score of $0.0$ indicates parity with the baseline, while a score of $1.0$ corresponds to consistently finding the global optimum immediately. 
The optimization budget per run is defined as the time required by the baseline to reach $95\%$ of the distance between the search space median and optimum. 
Each algorithm variant is executed $100$ times per search space to mitigate stochasticity.

\subsection{Impact of Target Application and Additional Information}
\label{subsec:evaluation_results_problem_specificity}

\begin{figure}[tb]
    \centering
    \includegraphics[width=\linewidth]{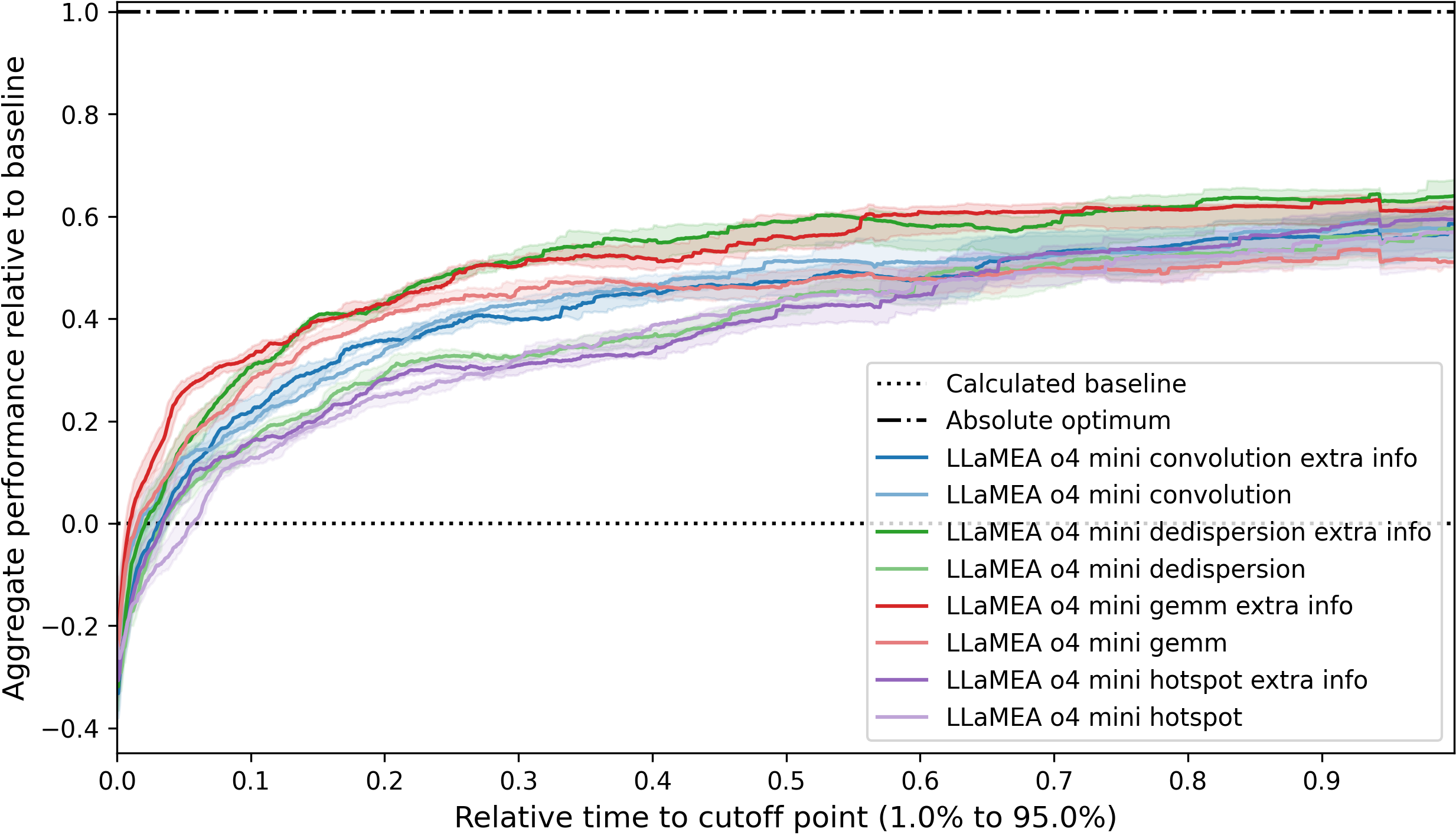}
    \caption{
    The aggregate performance over time over all search spaces of the application-specific optimization algorithms, each with and without additional search space knowledge. Mean of 100 runs each, the shaded area marks the 95\% confidence interval.}
    \label{fig:evaluation_aggregate_performance_extra_info}
\end{figure}

\begin{figure}[tb]
\centering
\subfigure[Convolution\label{fig:results_heatmap_convolution_no_info}]{
  \includegraphics[width=0.475\linewidth]{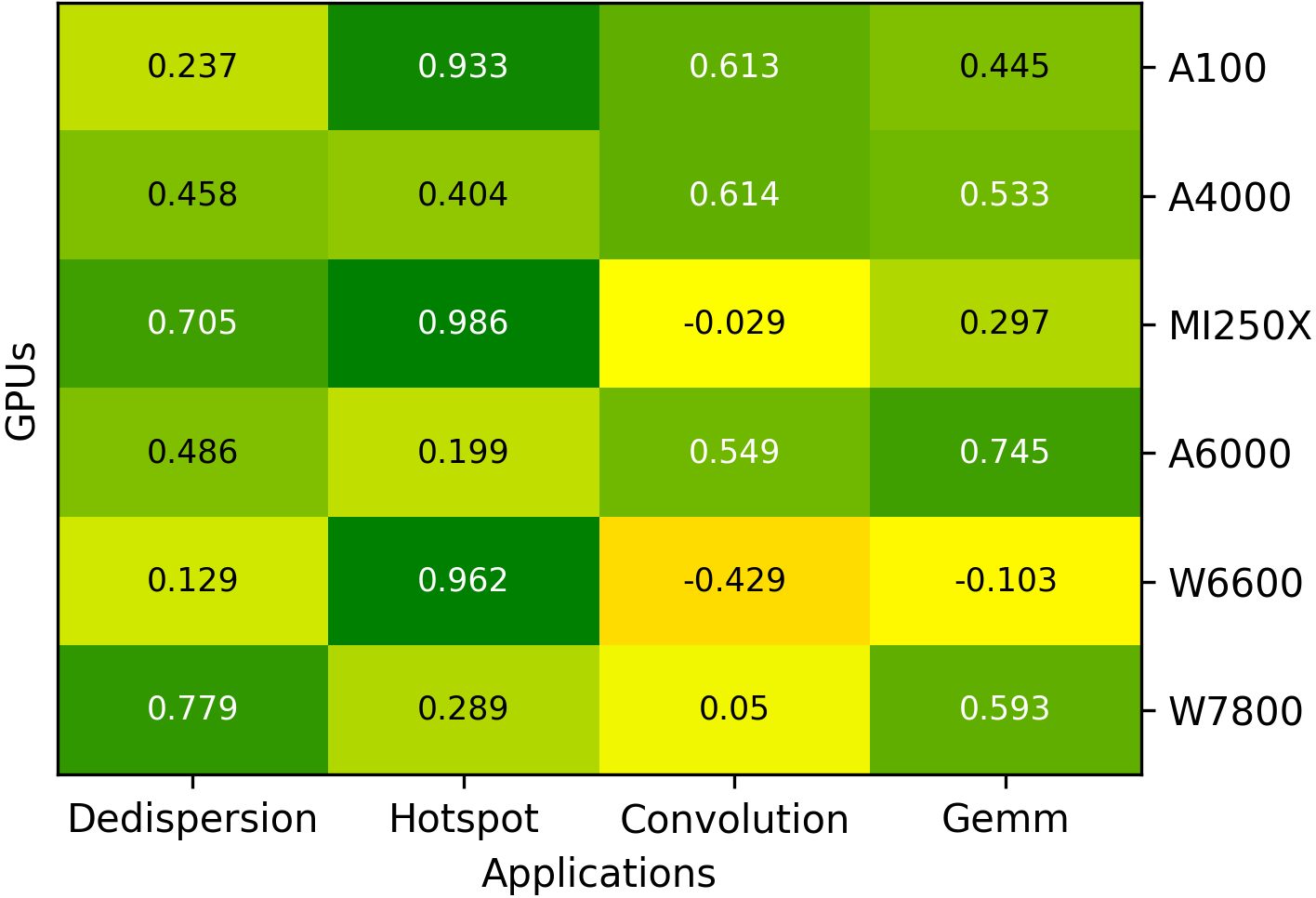}
}\hfil
\subfigure[Convolution with extra info\label{fig:results_heatmap_convolution_info}]{
  \includegraphics[width=0.475\linewidth]{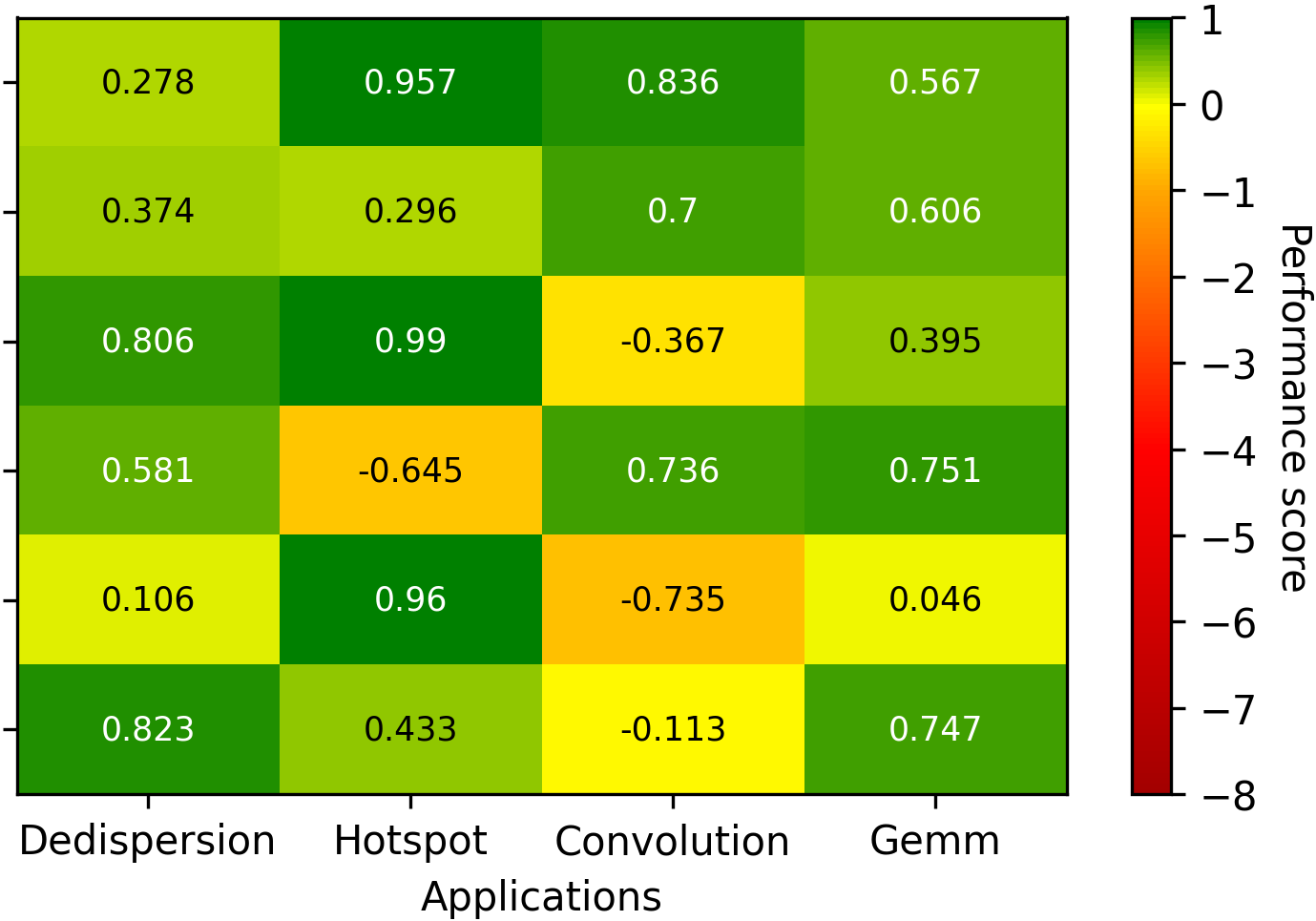}
} \\
\vspace{-3mm}
\subfigure[Dedispersion\label{fig:results_heatmap_dedispersion_no_info}]{
  \includegraphics[width=0.475\linewidth]{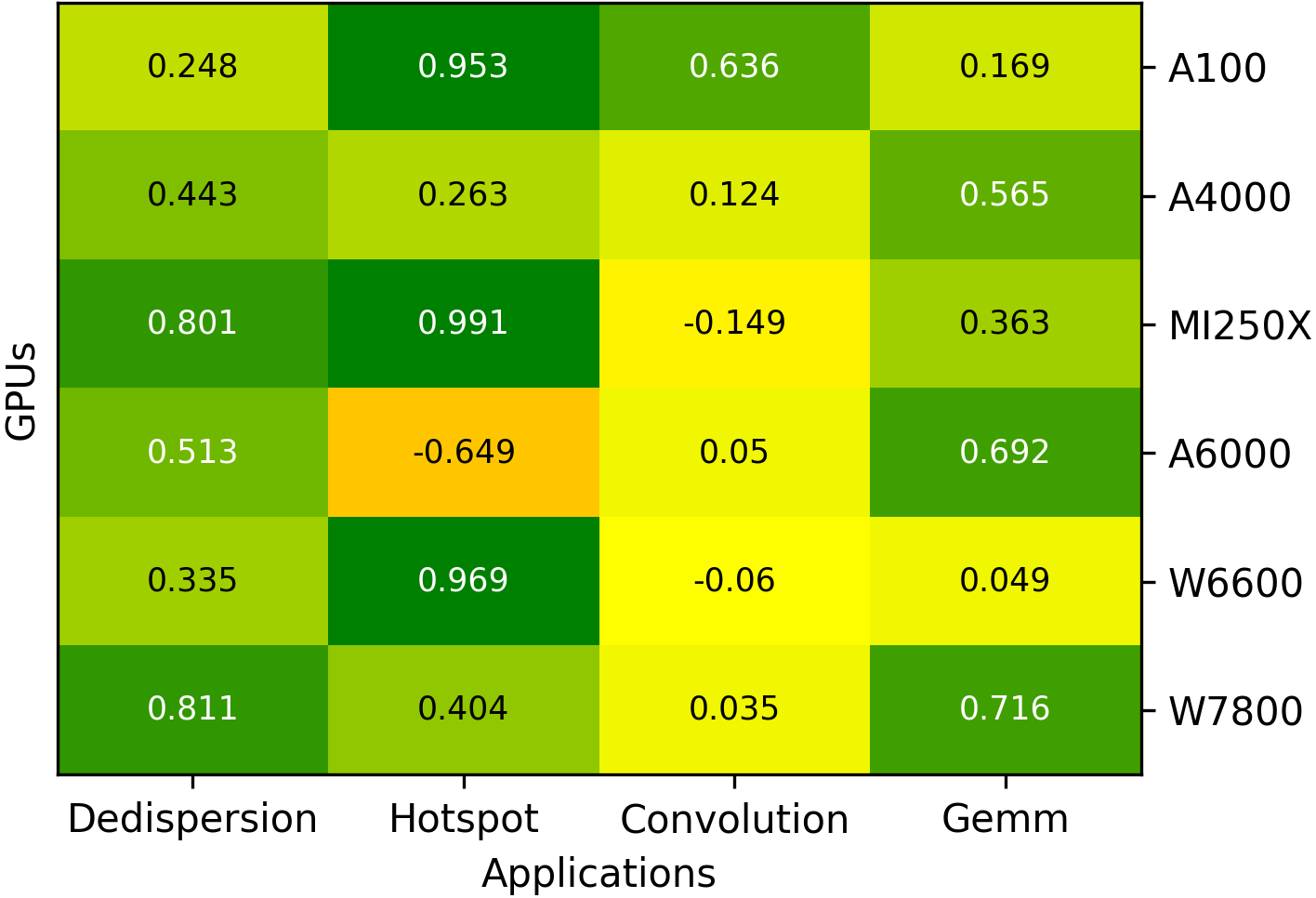}
}\hfil
\subfigure[Dedispersion with extra info\label{fig:results_heatmap_dedispersion_info}]{
  \includegraphics[width=0.475\linewidth]{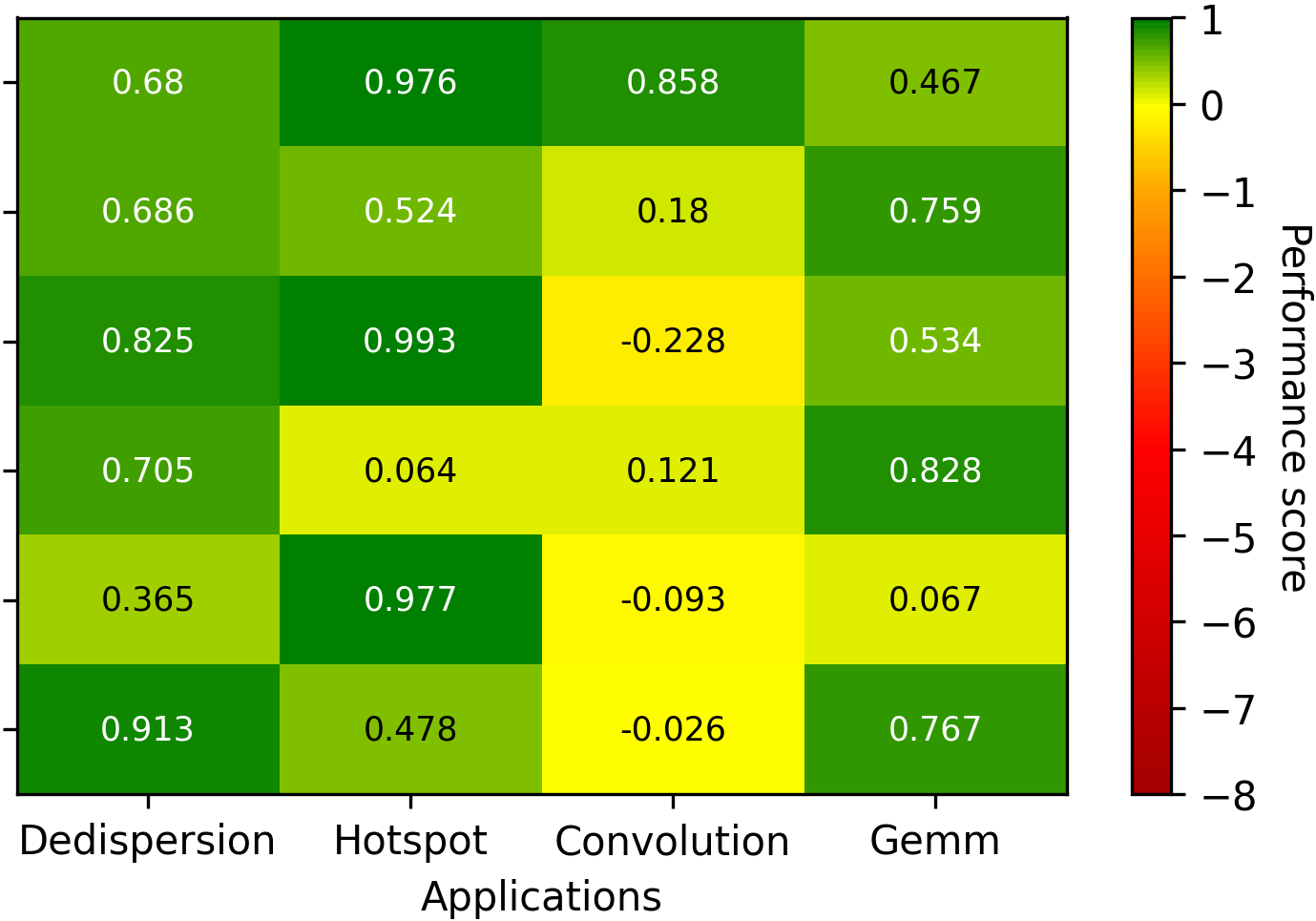}
} \\
\vspace{-3mm}
\subfigure[GEMM\label{fig:results_heatmap_gemm_no_info}]{
  \includegraphics[width=0.475\linewidth]{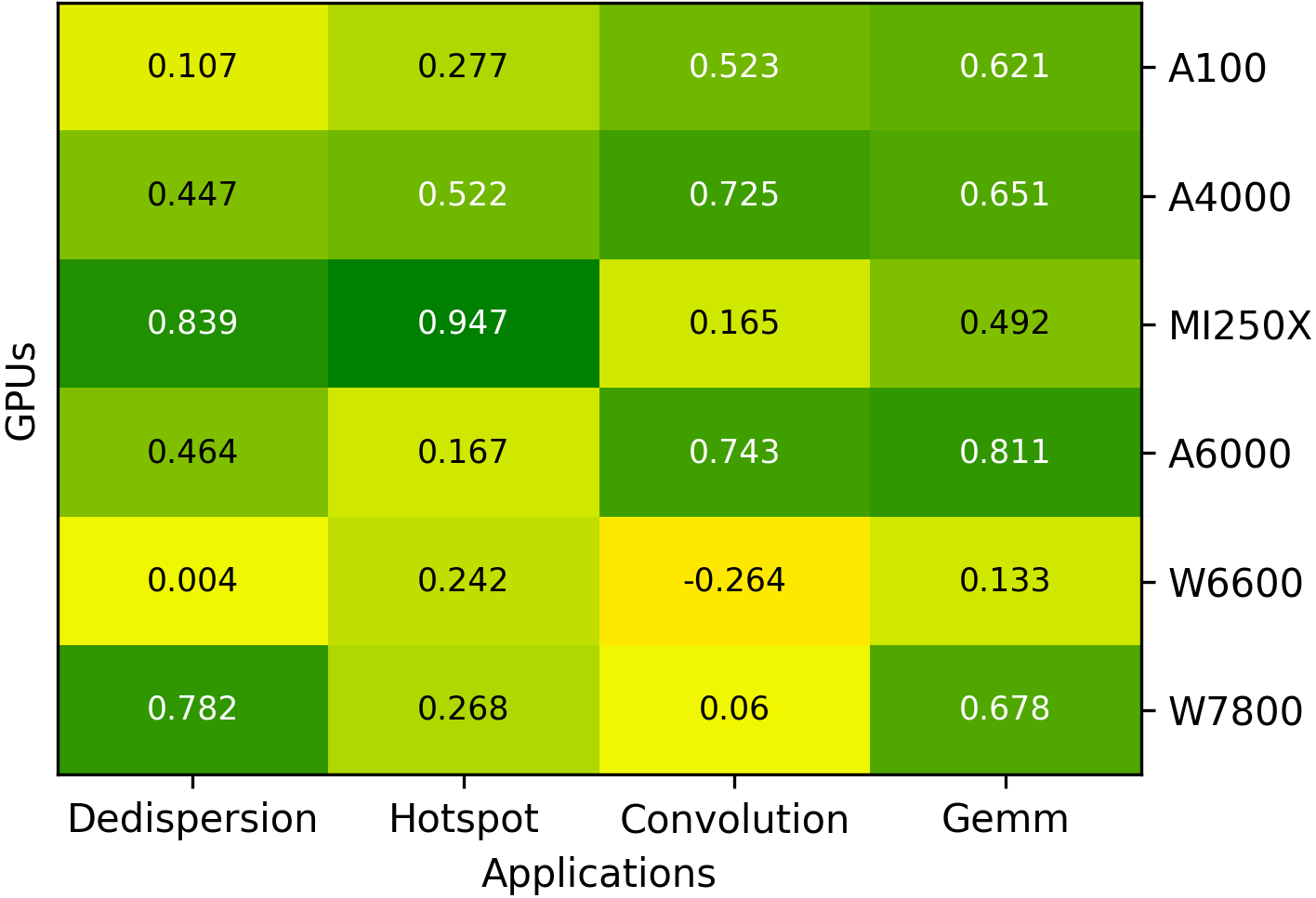}
}\hfil
\subfigure[GEMM with extra info\label{fig:results_heatmap_gemm_info}]{
  \includegraphics[width=0.475\linewidth]{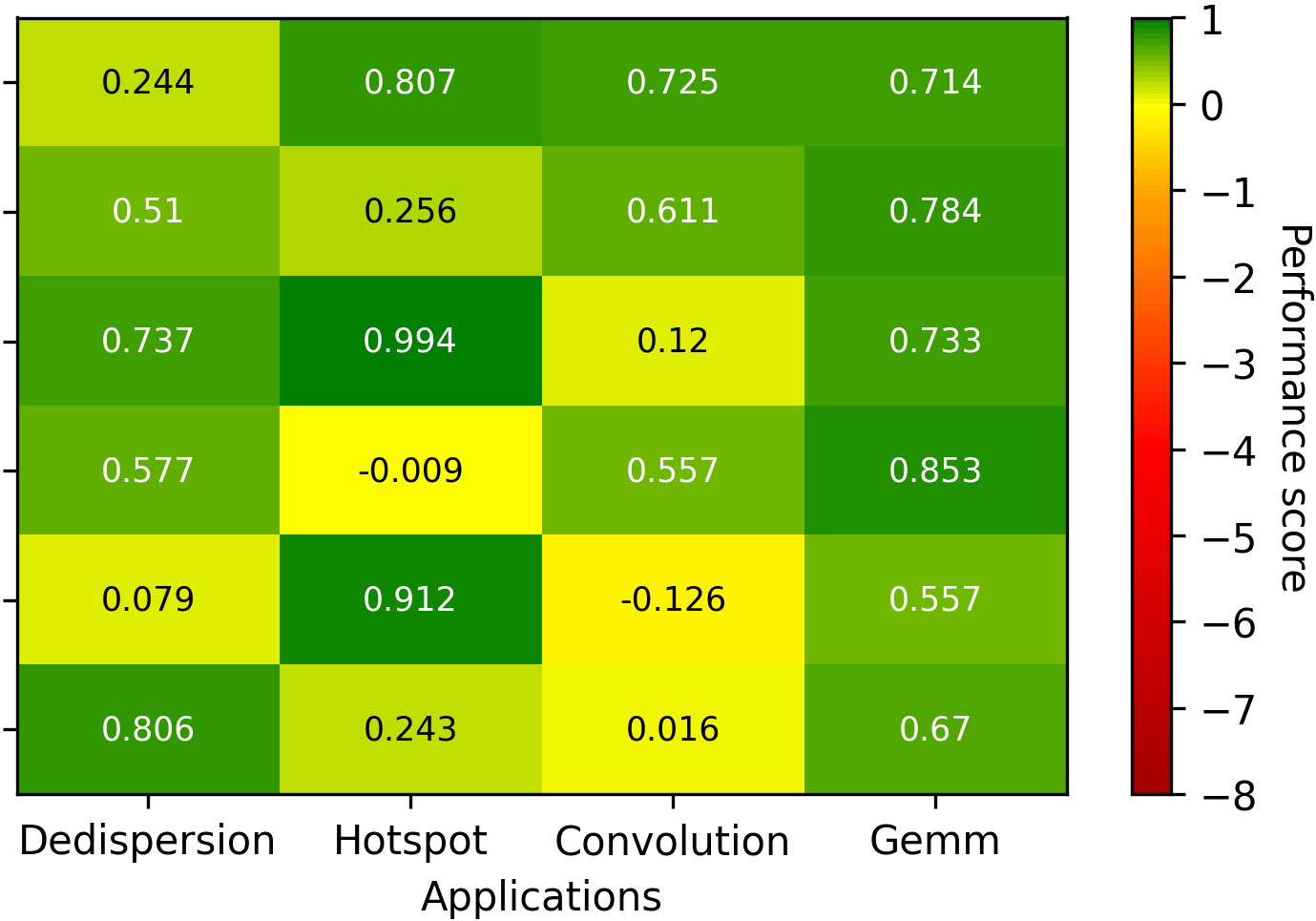}
} \\
\vspace{-3mm}
\subfigure[Hotspot\label{fig:results_heatmap_hotspot_no_info}]{
  \includegraphics[width=0.475\linewidth]{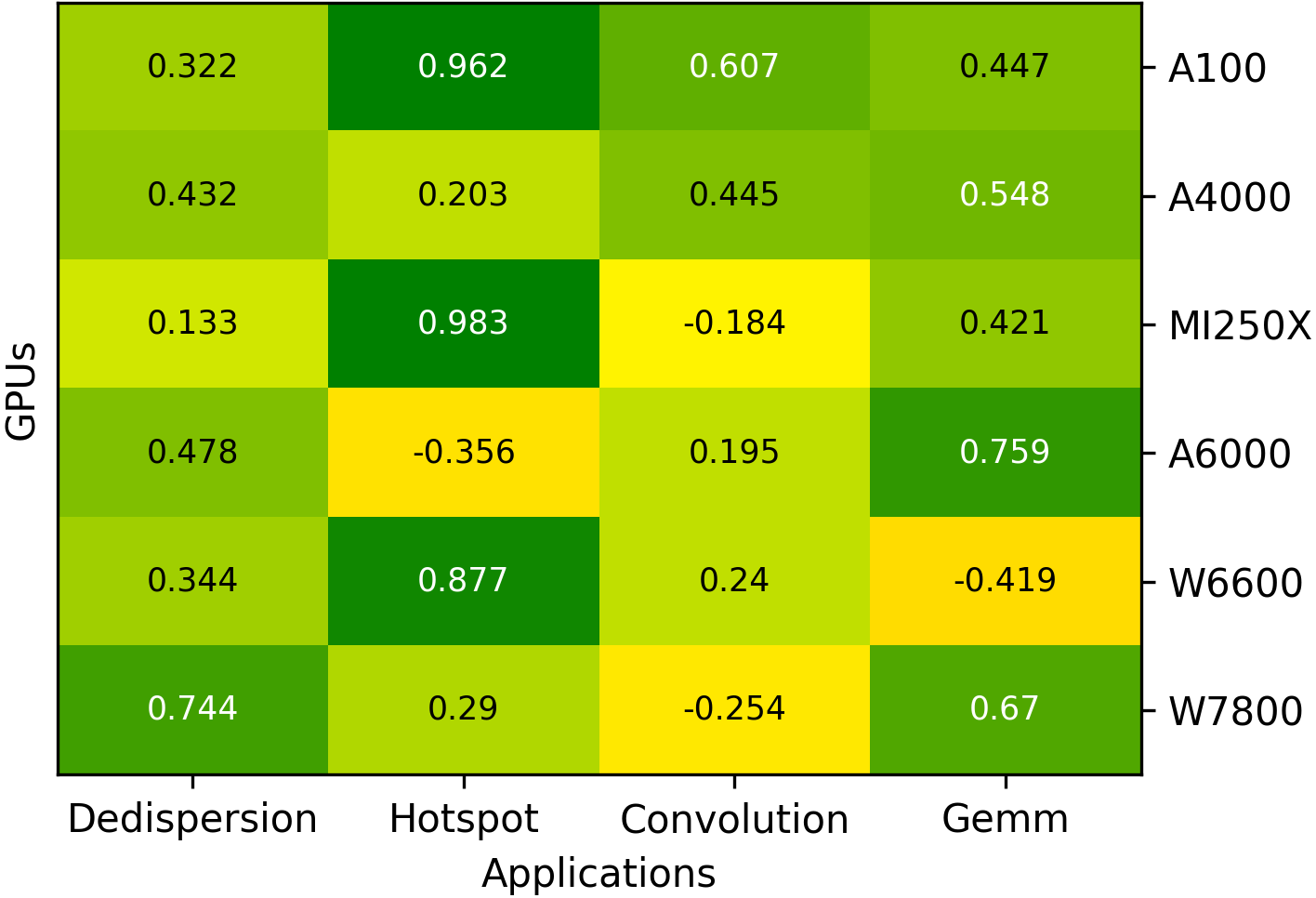}
}\hfil
\subfigure[Hotspot with extra info\label{fig:results_heatmap_hotspot_info}]{
  \includegraphics[width=0.475\linewidth]{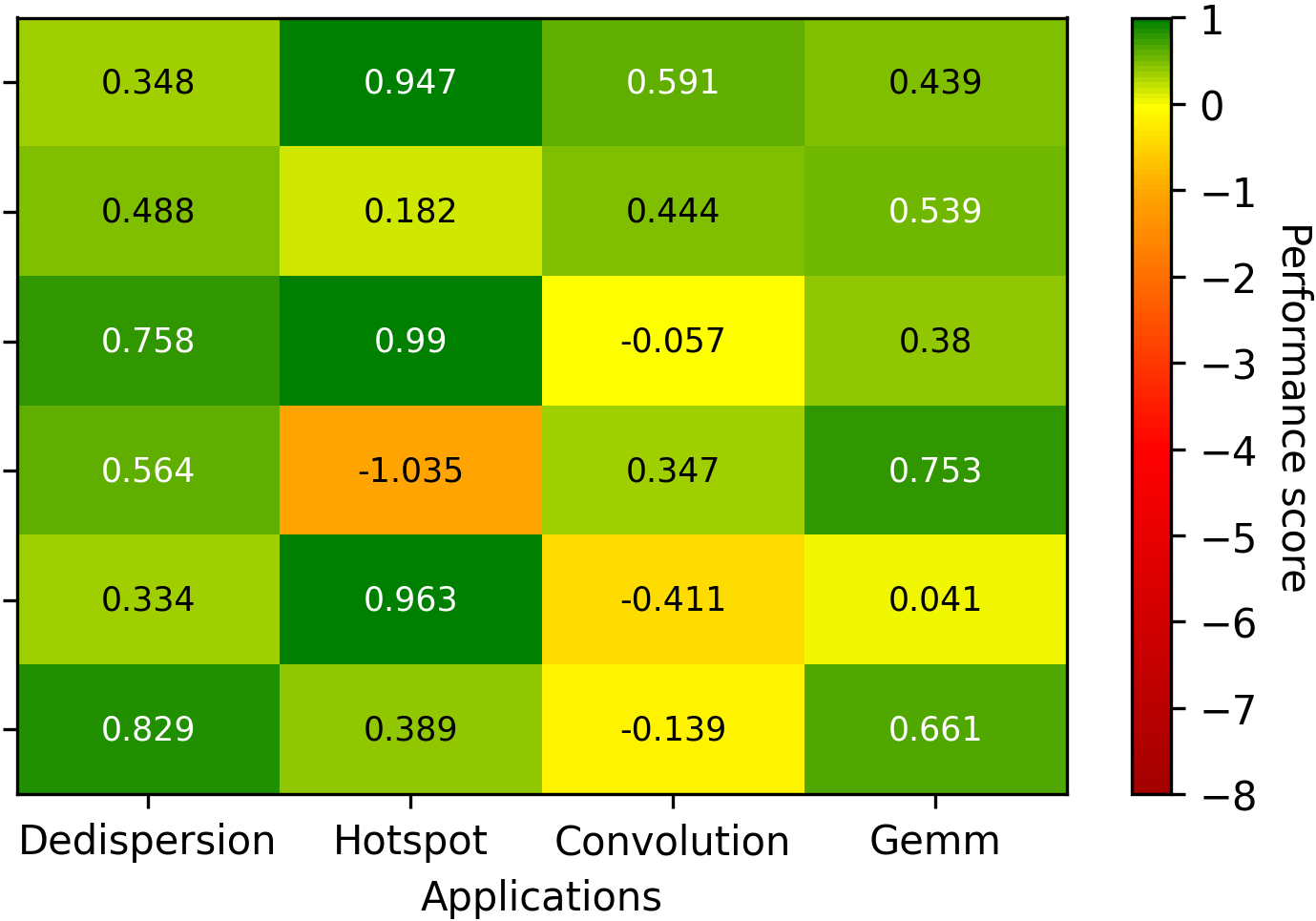}
}
\caption{Optimization algorithm performance per search space.}
\label{fig:results_heatmap_per_searchspace}
\end{figure}

For each application, we generated two algorithms using the LLaMEA-Kernel Tuner loop of \cref{sec:implementation} as per \cref{fig:prompt-template}:
\begin{enumerate}
    \item \textbf{Without search space information}: provided solely with an auto-tuning task description.
    \item \textbf{With search space information}: provided with information on possible parameter values and constraints. %
\end{enumerate}

Starting with the overall performance of both versions generated for each application across all search spaces in \cref{fig:evaluation_aggregate_performance_extra_info}, it is remarkable that all final versions of the generated optimization algorithms perform well in general. 
In these aggregate plots, each line denotes the mean performance over time of the 100 runs, and the shaded area its 95\% confidence interval. 
The performance scores are the means of this performance over time. 
As seen in \cref{tab:overall_performance_info_distinction}, the additional information on the search space provided in the prompt appears to have an overall positive influence, substantially so for the algorithms generated with target applications \textit{dedispersion} and \textit{GEMM}. On average, providing the additional information increased the performance score by $14.6\%$. 

\begin{table}[tb]
    \footnotesize
    \centering
    \caption{Overall performance scores and standard deviations of both algorithm versions for each target application.}
    \label{tab:overall_performance_info_distinction}
    \begin{tabularx}{\linewidth}{l|X|X|X}
    \toprule
        \textbf{Target application} & \textbf{Without extra info} & \textbf{With extra info} & \textbf{Difference} \\
    \midrule
        Convolution     & 0.429~\scriptsize{\textit{0.152}}     & 0.426~\scriptsize{\textit{0.157}}     & -0.003 \\
        Dedispersion    & 0.383~\scriptsize{\textit{0.162}}     & 0.515~\scriptsize{\textit{0.164}}     & +0.132 \\
        GEMM            & 0.432~\scriptsize{\textit{0.119}}     & 0.516~\scriptsize{\textit{0.136}}     & +0.084 \\
        Hotspot         & 0.373~\scriptsize{\textit{0.177}}     & 0.388~\scriptsize{\textit{0.172}}     & +0.015 \\
    \midrule
        \textit{Mean}   & 0.404     & 0.463     & +0.057 \\
    \bottomrule
    \end{tabularx}
\end{table}

\begin{table}[tbp]
    \footnotesize
    \centering
    \caption{Overall performance scores of non-target against target algorithms on the applications.}
    \label{tab:overall_performance_info_target_distinction}
    \begin{tabularx}{\linewidth}{l|X|X|X}
    \toprule
        \textbf{Target application} & \textbf{Non-target mean score} & \textbf{Target score} & \textbf{Difference} \\
    \midrule
        Convolution without extra info  & 0.195     & 0.219     & +0.024 \\
        Convolution with extra info     & 0.195     & 0.176     & -0.019 \\
        Dedispersion without extra info & 0.476     & 0.526     & +0.050 \\
        Dedispersion with extra info    & 0.476     & 0.695     & +0.219 \\
        GEMM without extra info         & 0.468     & 0.564     & +0.096 \\
        GEMM with extra info            & 0.468     & 0.719     & +0.251 \\
        Hotspot without extra info      & 0.538     & 0.493     & -0.045 \\
        Hotspot with extra info         & 0.538     & 0.404     & -0.134 \\
    \midrule
        \textit{Mean}                   & 0.419     & 0.475     & +0.055 \\
    \bottomrule
    \end{tabularx}
\end{table}

To further investigate these findings, we can compare the performance score per search space between the algorithms that did and did not receive additional search space information, shown in \cref{fig:results_heatmap_per_searchspace}. 
This performance score is the mean of the performance over time on each individual search space, as explained in \cref{subsec:integration}. 
This shows that while there is some correlation between the target application and performance, in general, the algorithms appear to generalize well, both outside their target application and outside the set of GPUs trained on. 
On a per-application level, we can distinguish between cases where an algorithm was targeted towards an application and where it was not. 
This is shown in \cref{tab:overall_performance_info_target_distinction}, where the \textit{non-target mean score} denotes the average performance score for that application of the algorithms not targeted towards it, which is compared to the two algorithms that were targeted for each application. 
While the two \textit{hotspot}-targeted algorithms and the \textit{convolution}-targeted algorithm with extra information perform worse than their non-targeted counterparts, the other five algorithms appear to have benefited substantially from the application-targeted approach, with an average improvement of $30.7\%$. 

\subsection{Algorithmic Details of the Two Best Generated Optimizers}
\label{subsec:evaluation_algorithm_details}

As reported in \cref{subsec:evaluation_results_problem_specificity}, enriching the prompt with search-space information particularly benefited the algorithms targeted at \emph{dedispersion} and \emph{GEMM}. 
To better understand the generated algorithms, we look at the algorithmic details and their approach to auto-tuning problems. 

\subsubsection{HybridVNDX (target Dedispersion, extra info)}
\label{sec:hybridvndx}

The first algorithm, \textit{HybridVNDX}, combines Variable Neighborhood Descent (VND)~\cite{Duarte2018} with (i) dynamic neighborhood weighting, (ii) a light k-NN surrogate for pre-screening, (iii) elite recombination, and (iv) tabu search \cite{gendreau2013tabu} and simulated-annealing \cite{van1987simulated}. 
The design aims to quickly filter promising neighbours per iteration while maintaining diversification and robust escape from local minima.
The default hyperparameters as used in this paper are: $k{=}5$, pool size $=8$, restart after $100$ non-improving steps, tabu size $=300$, elite size $=5$, $T_0{=}1.0$, cooling $=0.995$.

\begin{algorithm}[tb]
\label{alg:gemm_extra_info}
\caption{HybridVNDX high-level pseudocode}
\footnotesize
\DontPrintSemicolon
\SetAlgoNoEnd
\SetKwInOut{Input}{Input}\SetKwInOut{Output}{Output}
\Input{Objective $f$, search space $\mathcal{X}$ with neighbourhoods; budget via $f.\texttt{budget\_spent\_fraction}$}
\Output{Best configuration $x^\star$}
Initialize $x \leftarrow \texttt{random\_valid}()$, $f_x \leftarrow f(x)$; maintain history $\mathcal{H}$, elite heap $\mathcal{E}$, tabu deque $\mathcal{T}$;
Initialize neighbourhood weights $w[\cdot]\leftarrow 1$; temperature $T \leftarrow T_0$.

\While{$f.\texttt{budget\_spent\_fraction} < 1$}{
  Sample neighbourhood $N$ by roulette over $w$\; 
  
  Build candidate pool: subset of $N(x)$, 1 elite-crossover child, fill with random valid samples; repair infeasible\; 
  
  Score each candidate $c$ by k-NN prediction on $\mathcal{H}$ (Hamming), add tabu penalty; pick $\tilde{c}=\arg\min \text{score}$\; 
  
  Evaluate $f_{\tilde{c}} \leftarrow f(\tilde{c})$; push $(\tilde{c}, f_{\tilde{c}})$ to $\mathcal{H}$ and $\mathcal{E}$; \; 
  
  \lIf{$f_{\tilde{c}} \le f_x$ \textbf{or} $\text{rand}()<\exp(-(f_{\tilde{c}}-f_x)/T)$}{
     $x \leftarrow \tilde{c}$; $f_x \leftarrow f_{\tilde{c}}$; push $x$ to $\mathcal{T}$; $w[N]\!\leftarrow\!1.1w[N]$}
  \lElse{$w[N]\!\leftarrow\!0.9w[N]$}
  $T \leftarrow \alpha T$ (cooling);\; \If{stagnation $>$ threshold}{restart from new random valid $x$ and reset $T$}
}
Return the best-so-far in $\mathcal{H}$\;
\end{algorithm}

\subsubsection{AdaptiveTabuGreyWolf (target GEMM, extra info)}
\label{sec:leader-mixed-sa-tabu}

\begin{algorithm}[tb]
\caption{AdaptiveTabuGreyWolf}
\footnotesize
\DontPrintSemicolon
\SetAlgoNoEnd
\SetKwInOut{Input}{Input}\SetKwInOut{Output}{Output}
\Input{$f$, $\mathcal{X}$ with $v(\cdot)$ and $N_m(\cdot)$, budget $B$, hyperparameters $(p,L,s,q,\tau,\rho,T_0,\lambda,T_{\min})$}
\Output{$x^\star$}
$P \leftarrow$ $p$ random valid configs; evaluate; $\mathcal{T}\leftarrow$ recent list of size $L$; $x^\star\leftarrow \arg\min_{x\in P} f(x)$; $b\leftarrow 0$\;
\While{$b < 1$}{
  sort $P$ by $f$; set leaders $(\alpha,\beta,\delta)$ to the best three;\;
  \ForEach{$x\in P\setminus\{\alpha,\beta,\delta\}$}{
    \tcp{Leader-mixed proposal}
    $y_j \leftarrow \text{Uniform}\{ \alpha_j,\beta_j,\delta_j,x_j\}$ for each $j$;\;
    \tcp{Shaking}
    with prob.\ $s$: \lIf{rand() $< q$}{random-dim jump from random valid sample} \lElse{$y\leftarrow$ one-step move in $N_{m(b)}(y)$}\;
    \tcp{Repair, tabu}
    \If{$\neg v(y)$}{attempt repair via neighbors; else resample random valid}\;
    \If{$y\in\mathcal{T}$}{resample (small Hamming change or fresh sample)}\;
    \tcp{Evaluate and accept}
    Evaluate $f(y)$; $\Delta\leftarrow f(y)-f(x)$; $T\leftarrow \max(T_{\min},T_0 e^{-\lambda b})$\;
    \If{$\Delta \le 0$ \textbf{or} rand() $< e^{-\Delta/T}$}{$x\leftarrow y$; push $x$ to $\mathcal{T}$}\;
  }
  update $x^\star$ and stagnation counter; \If{stagnation $\ge \tau$}{reinit worst $\rho p$ individuals}\;
  update budget fraction $b$;
}
\Return{$x^\star$}
\end{algorithm}

The second algorithm, \textit{AdaptiveTabuGreyWolf}, keeps a small population of valid configurations and, at each step, proposes new candidates for every non-leader by mixing each parameter independently from the three current best solutions or the individual itself. 
A light ``shaking'' step then perturbs the proposal either by a random coordinate jump from a fresh valid sample or by a one-step move in a discrete neighborhood (using coarser adjacent moves early and stricter ones later) to balance exploration and exploitation. Infeasible proposals are repaired; recently evaluated points are blocked by a tabu list to avoid repeats. 
Candidates are accepted with simulated-annealing criteria under a temperature that decays with budget (with mild reheating on stagnation), and if progress stalls, a fraction of the worst population members is randomly reinitialized. 
The run stops when the evaluation budget is exhausted and the best-so-far configuration is returned.
While this algorithm is named differently, after close inspection, various elements are similar to the first (e.g., the tabu list and simulated annealing approach). 
The default hyperparameters as used in this paper are: Population size $p{=}8$, tabu length $L{=}3p$, shake rate $s{=}0.2$, jump rate $q{=}0.15$, stagnation limit $\tau{=}80$, restart ratio $\rho{=}0.3$, $T_0{=}1.0$, $\lambda{=}5.0$, $T_{\min}{=}10^{-4}$.

These two variants provided the best aggregate performance across the $24$ search spaces. 
To better understand what optimization algorithms perform well on auto-tuning problems, we conclude that they share two important characteristics: 
1) Both methods treat evaluation time as dominant; their additional control logic is lightweight. 
2) The neighbour APIs allow architecture-agnostic moves while honouring constraints, which is crucial for large irregular search spaces.

\subsection{Comparison to Human-designed Auto-Tuning Algorithms}
\label{subsec:evaluation_results_traditional}
Having evaluated the quality and details of generated optimization algorithms in \cref{subsec:evaluation_results_problem_specificity,subsec:evaluation_algorithm_details}, we now evaluate how they hold up against human-designed auto-tuning optimization algorithms. 
We compare the best two of our generated algorithms, those targeted towards \textit{dedispersion} and \textit{GEMM} with extra information, against the two best human-designed auto-tuning methods in Kernel Tuner evaluated in \citet{willemsenTuningTheTuner2025}. 
These are Genetic Algorithm (GA), a population-based evolutionary method, and Simulated Annealing (SA), a local search technique with probabilistic acceptance. %
Both algorithms have undergone extensive hyperparameter tuning for 7 days on the same search spaces as used during the generation stage in this work, under the same conditions, further specified in~\citet{willemsenTuningTheTuner2025}. 
To further expand the context of this evaluation, we also compare against the best-performing optimization algorithm from another auto-tuning framework, pyATF~\cite{pyATF}, which uses a Differential Evolution (DE) implementation. 
Hyperparameter tuning of pyATF optimizers is not possible without changing the source code. %
The pyATF version used is 0.0.9. 

We have chosen these methods as our comparison baselines due to their prevalence in Kernel Tuner and similar frameworks, representing a dominant class of human-designed optimizers. While other ML-based auto-tuners, such as GPTune and HyperMapper, are potentially relevant baselines, they are not directly comparable in this evaluation due to differing device targets and underlying assumptions regarding continuous spaces and compiler-internal cost models.

\begin{figure}[tb]
    \centering
    \includegraphics[width=\linewidth]{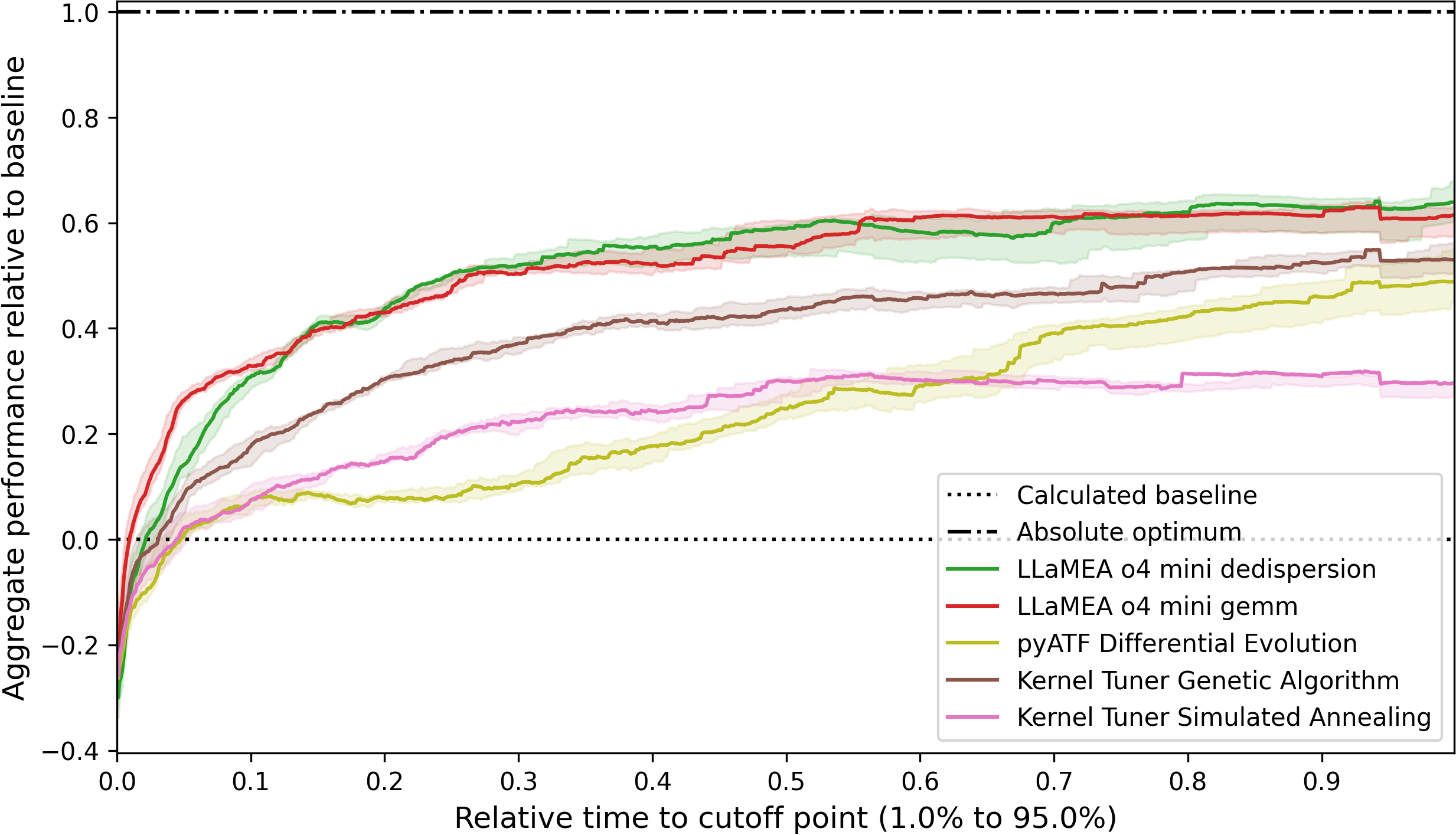}
    \caption{
    The aggregate performance over time over all search spaces of our generated algorithms against three human-designed optimization algorithms. Mean of 100 runs each, the shaded area marks the 95\% confidence interval.}
    \label{fig:evaluation_aggregate_performance_against_kt}
\end{figure}

\begin{figure}[tb]
\centering
\subfigure[Dedispersion with extra info\label{fig:results_heatmap_compare_kt_dedispersion_info}]{
  \includegraphics[width=0.475\linewidth]{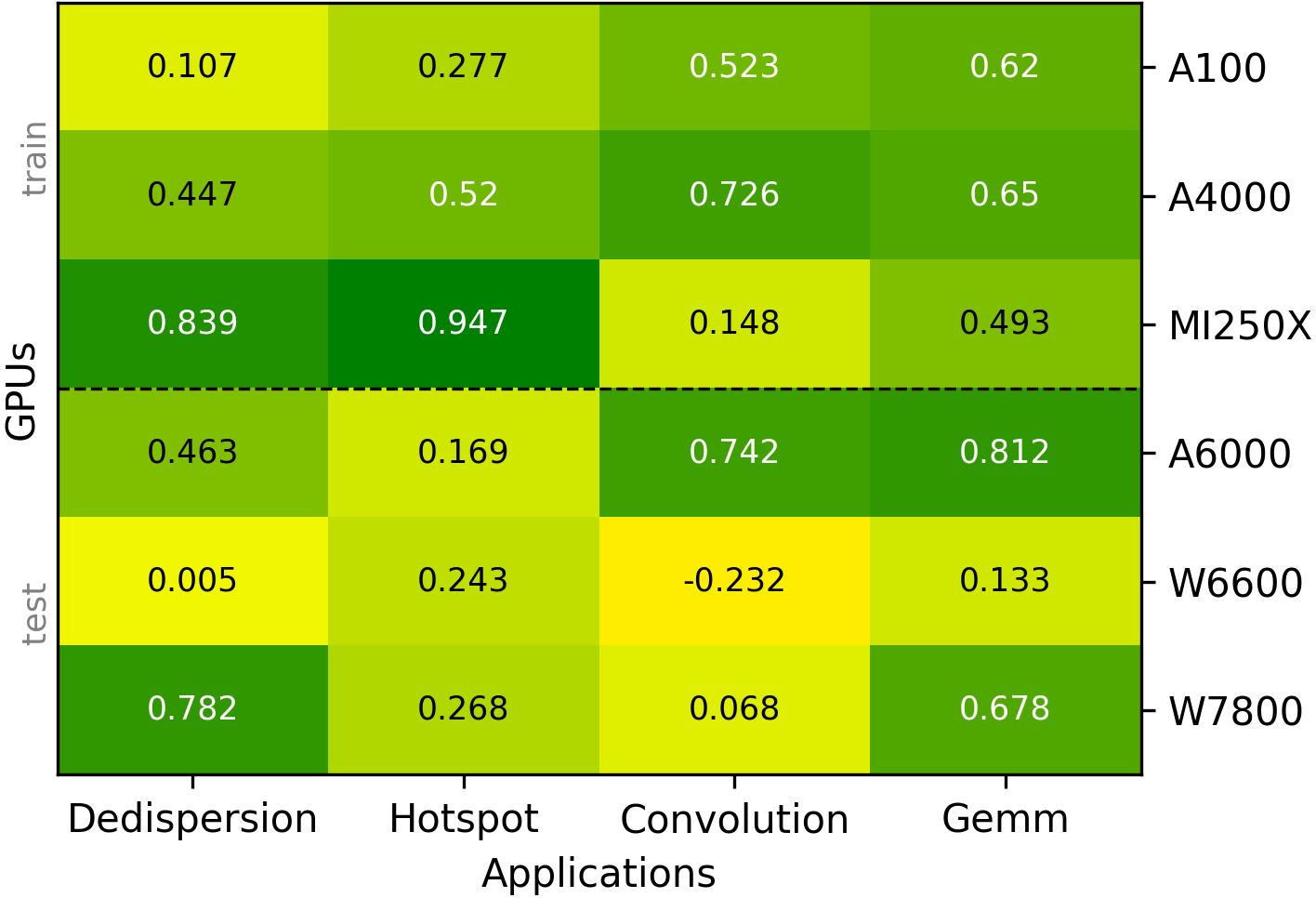}
}\hfil
\subfigure[GEMM with extra info\label{fig:results_heatmap_compare_kt_gemm_info}]{
  \includegraphics[width=0.475\linewidth]{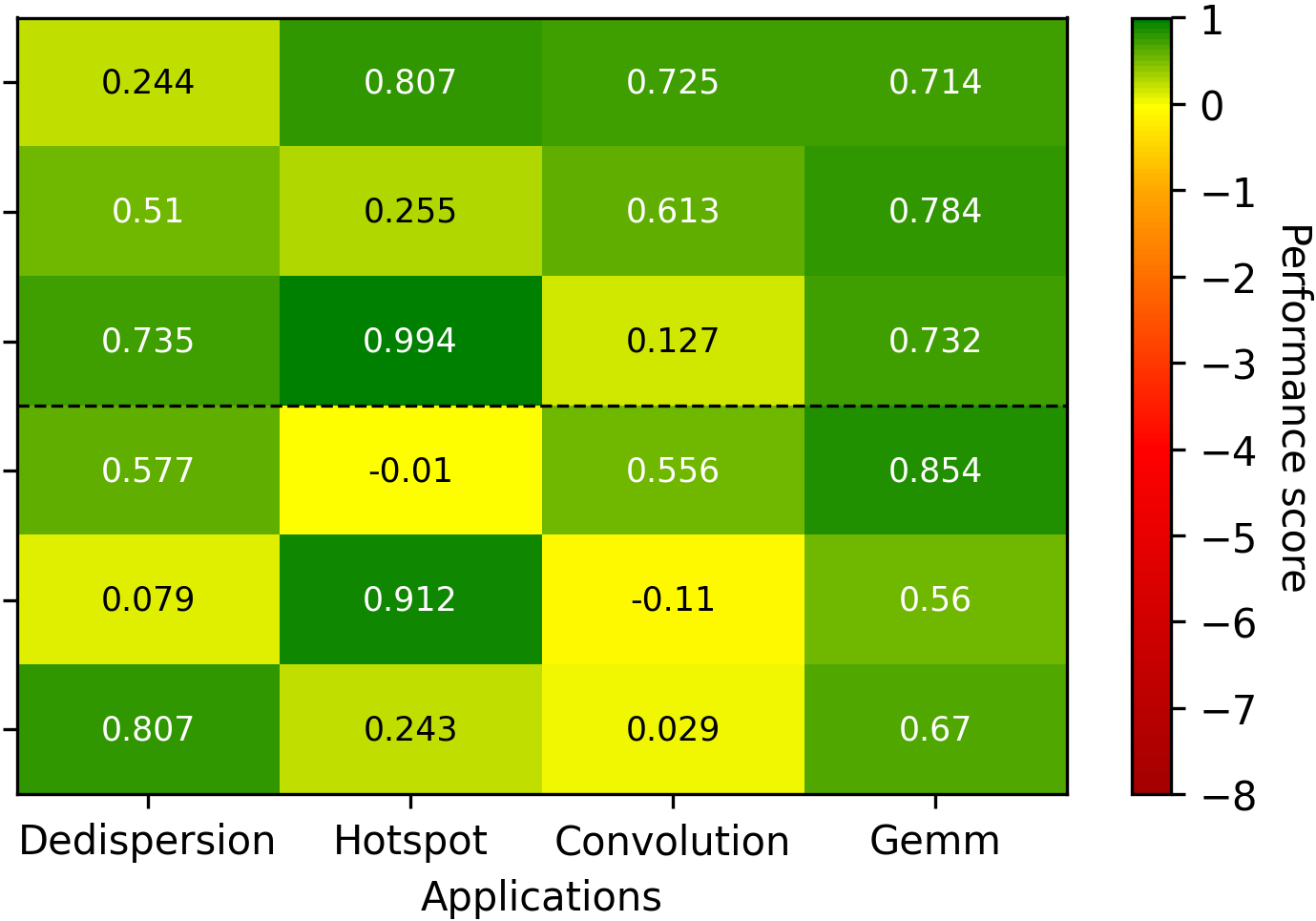}
} \\
\vspace{-3mm}
\subfigure[Kernel Tuner GA\label{fig:results_heatmap_compare_kt_kt_ga}]{
  \includegraphics[width=0.475\linewidth]{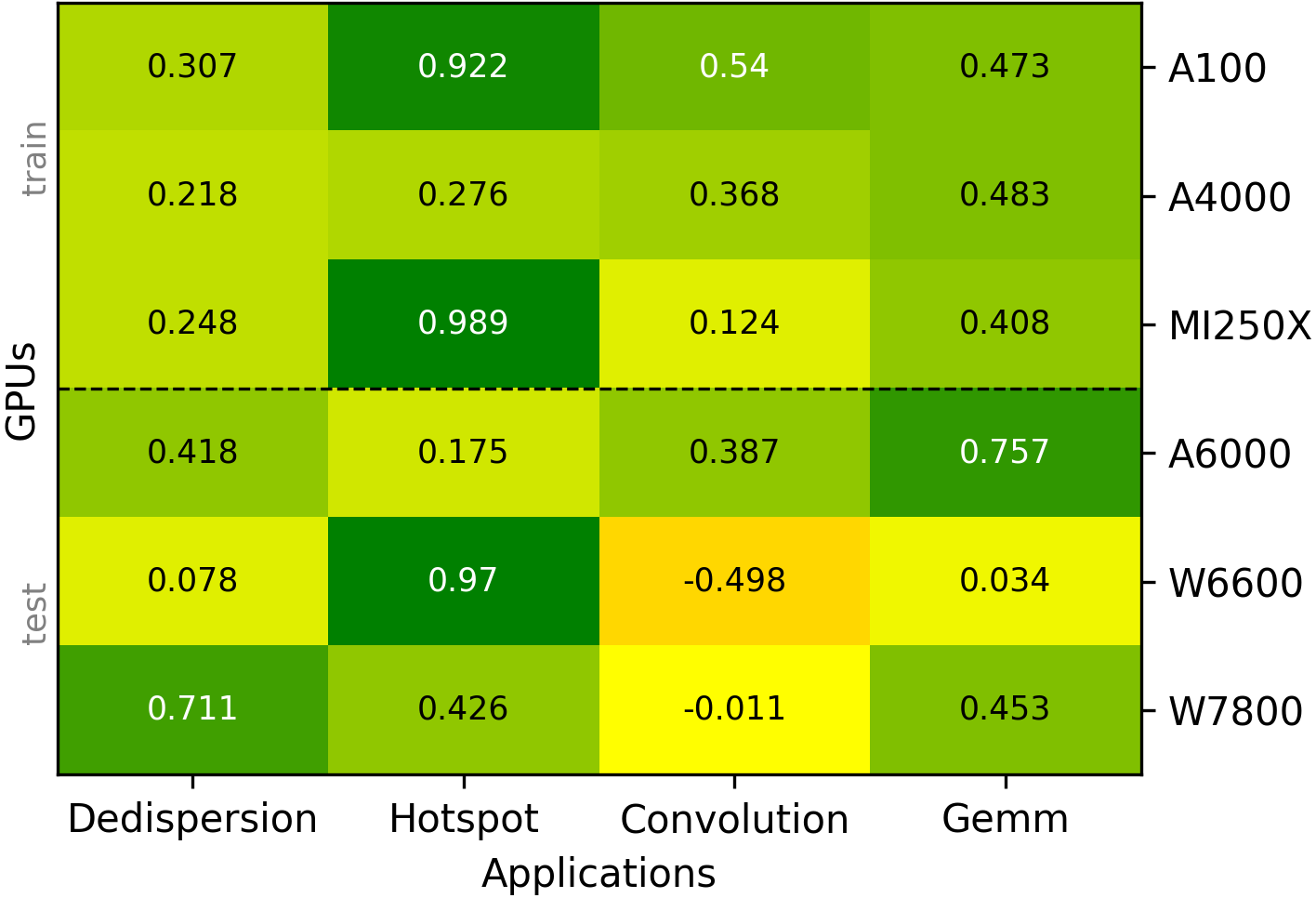}
}\hfil
\subfigure[Kernel Tuner SA\label{fig:results_heatmap_compare_kt_kt_sa}]{
  \includegraphics[width=0.475\linewidth]{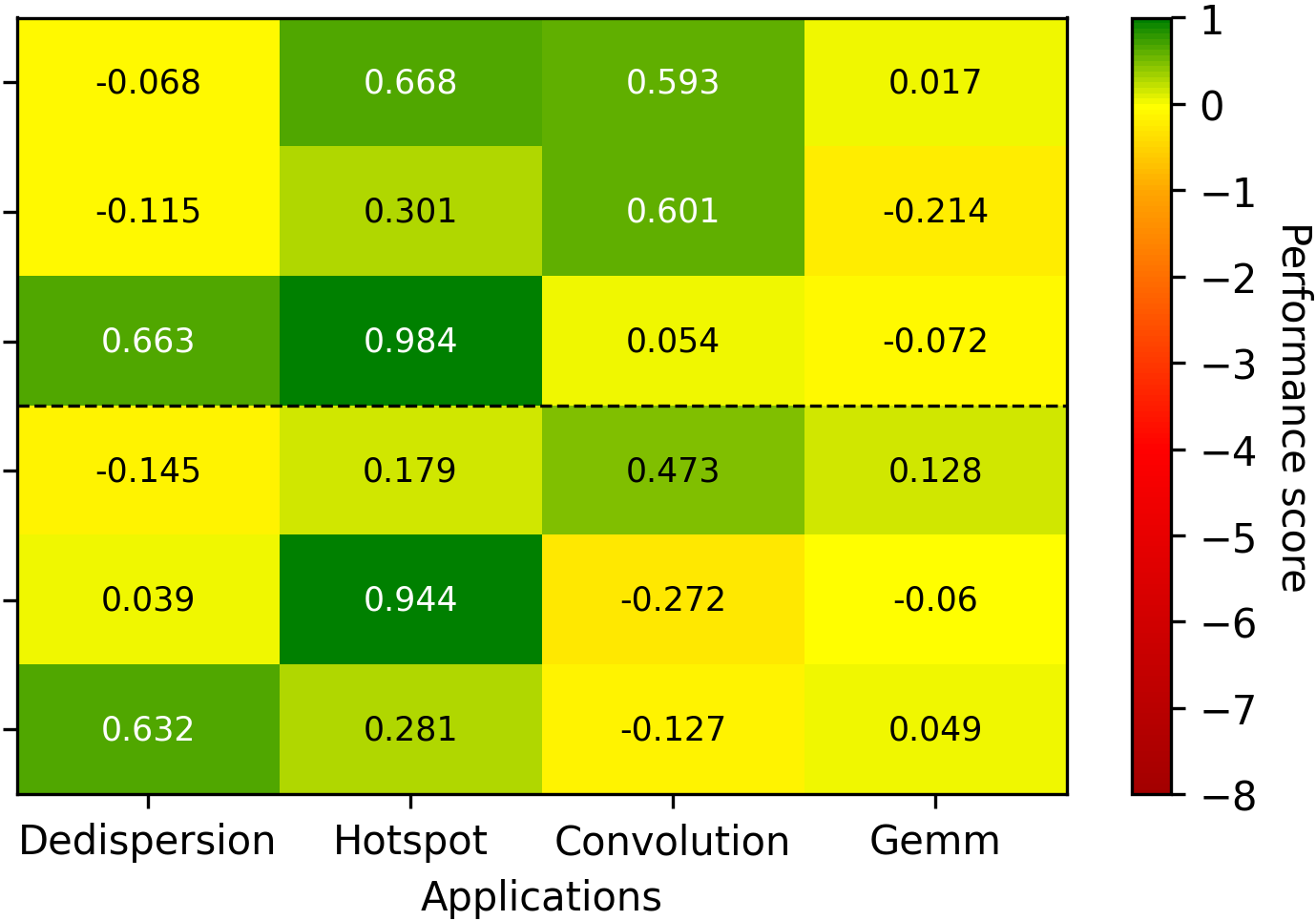}
} \\
\vspace{-3mm}
\subfigure[pyATF DE\label{fig:results_heatmap_compare_kt_pyatf_de}]{
  \includegraphics[width=0.475\linewidth]{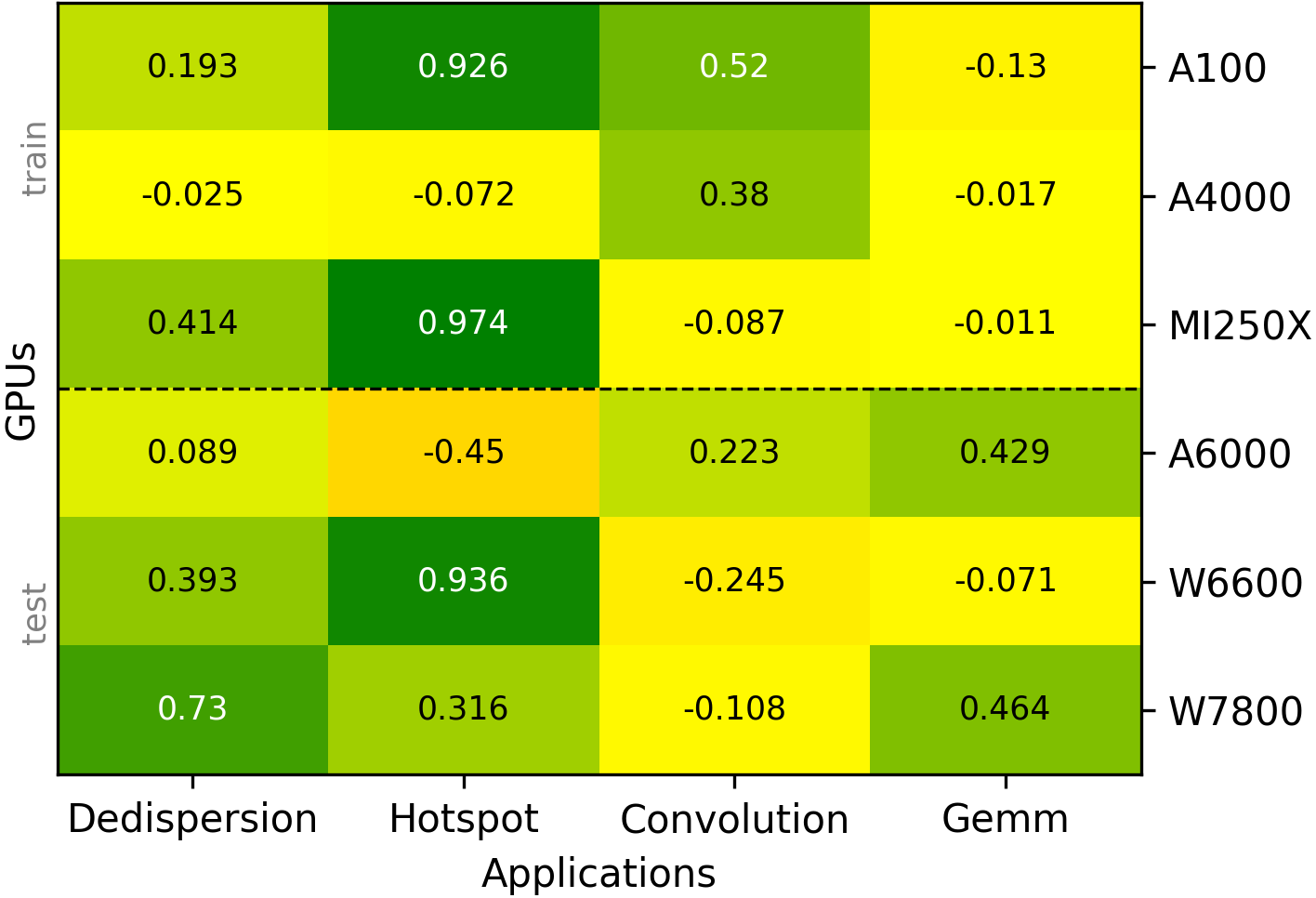}
}
\caption{Optimization algorithm performance per search space.}
\label{fig:results_heatmap_per_searchspace_against_kt}
\end{figure}

\Cref{fig:evaluation_aggregate_performance_against_kt} shows the aggregate performance of the two LLM-generated algorithms and the three human-designed optimization algorithms across all 24 search spaces. 
The results indicate that our LLM-generated algorithms achieve more than competitive performance as they outperform the human-designed optimization algorithms by a substantial margin, improving the performance score by 0.126 and 0.282 over Kernel Tuner's GA and SA, respectively. 
Over pyATF's DE, the performance score is improved by 0.274. 
On average, our LLM-generated algorithms perform $72.4\%$ better than the human-designed algorithms compared to across the board. 
Performance stability across search spaces is also notable, as seen in \cref{fig:results_heatmap_per_searchspace_against_kt}. 
Comparing the performance of our generated algorithms against the best human-designed algorithm in the evaluation, GA, it is notable that the latter performs better on the \textit{hotspot} application, but is otherwise outperformed in general.

Overall, these results demonstrate that while LLMs can already synthesize competitive optimization algorithms in a general setting, providing additional problem-specific information further boosts their performance, enabling them to surpass human-designed optimization methods. 

\section{Conclusion}
\label{sec:conclusion_futurework}

As modern architectures continue to grow in complexity, auto-tuning remains indispensable for achieving high performance. %
At the heart of auto-tuning lies the choice of optimization algorithm, which governs how efficiently the vast, irregular search spaces can be explored. In this work, we investigated a novel approach: leveraging large language models (LLMs) to automatically generate optimization algorithms tailored to auto-tuning problems. 
While components of our framework existed independently, their integration into a closed-loop meta-optimization system produces emergent behavior: the discovery of novel optimization algorithms not present in either component. 

Our results show that LLM-generated optimization algorithms can achieve performance competitive with, and in various cases superior to, established human-designed algorithms. 
We observed that these auto-generated methods adapt well to diverse tuning spaces, demonstrating both strong search efficiency over time and the ability to discover high-performing configurations with limited evaluations.
Our results show that providing application- and search-space-specific information leads to substantial improvements, highlighting the value of specialization. At the same time, many auto-tuning problems share common structural characteristics, allowing some optimizers generated for one problem to transfer effectively to other search spaces. The strong performance observed across diverse tasks therefore does not contradict specialization; rather, it indicates that our automated design approach can uncover principled optimization strategies that exploit shared structure, while still benefiting from additional problem-specific information when available. 
This approach opens new avenues for rapid design and exploration of optimization strategies without requiring extensive manual algorithm engineering.

Building on these promising results, we identify several directions for future work.
First, the generated algorithms and their performance depend on the choice of the underlying LLM. 
We experimented with various models, including \texttt{o4-mini} and \texttt{gemini-2.0-flash}. %
Although \texttt{gemini-2.0-flash} produced valid algorithms, their performance was generally inferior to those generated by \texttt{o4-mini}. 
We selected \texttt{o4-mini} for the presented experiments because it offered a favorable balance between performance and computational cost. 
In addition, while we demonstrate that LLM-generated algorithms can generalize to a diverse set of unseen search spaces, our study relies on a fixed set of applications and GPUs to evaluate generated algorithms. Broader validation across additional architectures and domains can provide further details on their robustness. 
Finally, this work is a first step into a new realm, where optimizer generation could become a foundational capability within large-scale auto-tuning infrastructures, compilers, and ML deployment systems. By automatically producing specialized optimizers, such systems could achieve unprecedented levels of portability and self-adaptation across diverse hardware and workloads.

In conclusion, our study demonstrates the potential of LLMs to reshape the landscape of auto-tuning by automating the design of optimization algorithms themselves.
This paradigm shift not only reduces the human effort required to craft effective search strategies but also paves the way toward more flexible, efficient, and accessible auto-tuning for the next generation of high-performance computing systems.
\ifdoubleblind
\else
The best-performing optimization algorithms in this work are available in Kernel Tuner, which can be installed with \verb|pip install kernel-tuner|. 
\fi
\ifdoubleblind
\else
LLaMEA can be installed with \verb|pip install llamea|. 
The complete experiment is implemented using the BLADE \cite{van_Stein_BLADE_Benchmark_suite_2025} benchmarking suite, and can be found in our \href{https://github.com/XAI-liacs/BLADE}{Github repository}\footnote{\tiny\url{https://github.com/XAI-liacs/BLADE/tree/paper/auto-tuning}}.
\fi
\ifdoubleblind
\else
For more information, please visit the \href{https://github.com/KernelTuner/kernel_tuner}{Kernel Tuner}\footnote{\tiny\url{https://github.com/KernelTuner/kernel_tuner}} and \href{https://github.com/XAI-liacs/LLaMEA}{LLaMEA}\footnote{\tiny\url{https://github.com/XAI-liacs/LLaMEA}} repositories. 
\fi

\bibliography{references}
\bibliographystyle{mlsys2025}

\end{document}